\documentclass{article}
\usepackage{amssymb}
\usepackage{subcaption}
\usepackage{graphicx}
\usepackage[T1]{fontenc}
\usepackage{pgffor}
\usepackage{amssymb}
\usepackage{pifont}
\usepackage{xcolor}
\usepackage{wrapfig}

\newcommand{\cmark}{\ding{51}}%
\newcommand{\xmark}{\ding{55}}%
\newcommand{\frameworkName}[0]{JaxRobotarium}

\newcounter{myenum}
\newenvironment{flushenumerate}{%
 \begin{list}{\arabic{myenum}.}%
   {\setlength{\leftmargin}{15pt}}%
    \setlength{\labelwidth}{20pt}
    \setlength{\itemindent}{0pt}
    \setlength{\labelsep}{0.5em}
 \setlength{\itemsep}{1pt}
 \setlength{\parskip}{0pt}
 \setlength{\parsep}{0pt}
    \usecounter{myenum}}%
{\end{list}}

\newenvironment{flushitemize}{%
\begin{list}{$\bullet$}
   {\setlength{\leftmargin}{15pt}}%
    \setlength{\labelwidth}{20pt}
    \setlength{\itemindent}{0pt}
    \setlength{\labelsep}{0.5em}
 \setlength{\itemsep}{1pt}
 \setlength{\parskip}{0pt}
 \setlength{\parsep}{0pt}}
{\end{list}}

\usepackage[final]{corl_2025} % Uncomment for pre-prints (e.g., arxiv); This is like ``final'', but will remove the CORL footnote.

\title{\frameworkName: Training and Deploying \\ Multi-Robot Policies in 10 Minutes}

\author{
  Shalin Anand Jain, Jiazhen Liu, Siva Kailas, Harish Ravichandar\\
  Georgia Institute of Technology, 
  United States\\
}

\begin{document}
\maketitle

%===============================================================================

\begin{abstract}
    Multi-agent reinforcement learning (MARL) has emerged as a promising solution for learning complex and scalable coordination behaviors in multi-robot systems. However, established MARL platforms (e.g., SMAC and MPE) lack robotics relevance and hardware deployment, leaving multi-robot learning researchers to develop bespoke environments and hardware testbeds dedicated to the development and evaluation of their individual contributions. 
    The Multi-Agent RL Benchmark and Learning Environment for the Robotarium (MARBLER) is an exciting recent step in providing a standardized robotics-relevant platform for MARL, by bridging the Robotarium testbed with existing MARL software infrastructure. 
    % While MARBLER enables the deployment of trained polices on an open-source no-cost hardware test bed, it
    However, MARBLER lacks support for parallelization and GPU/TPU execution, making the platform prohibitively slow compared to modern MARL environments and hindering adoption. 
    We contribute \frameworkName, a  Jax-powered end-to-end simulation, learning, deployment, and benchmarking platform for the Robotarium. \frameworkName{} enables rapid training and deployment of multi-robot reinforcement learning (MRRL) policies with realistic robot dynamics and safety constraints, supporting both parallelization and hardware acceleration. Our generalizable learning interface provides an easy-to-use integration with SOTA MARL libraries (e.g., JaxMARL). 
    In addition, \frameworkName{} includes eight standardized coordination scenarios, including four novel scenarios that bring established MARL benchmark tasks (e.g., RWARE and Level-Based Foraging) to a realistic robotics setting. We demonstrate that \frameworkName{} retains high simulation fidelity while achieving dramatic speedups over baseline (20x in training and 150x in simulation), and provides an open-access sim2real evaluation pipeline through the Robotarium testbed, accelerating and democratizing access to multi-robot learning research and evaluation. Our code is publicly available on \texttt{\href{https://star-lab.cc.gatech.edu/papers/jain-jaxrobotarium/}{github}}.
\end{abstract}


%===============================================================================
\section{Introduction}
%%% Wide applications of MRS, how its time complexity induces interest in MRRL 
% Multi-robot systems have attracted increasing research attention for their wide potential applications in warehouse automation~\cite{8956244}, agriculture~\cite{mrs_agriculture}, surveillance~\cite{mrs_surveillance}, and search and rescue~\cite{queralta2020collaborative}. Central challenges in these domains often exhibit exponential time complexity as the number of robots increases or require complex coordination behaviors that are hard to specify in a generalizable manner. These issues underscore the need for efficient and adaptable coordination strategies. In this context, Multi-Robot Reinforcement Learning (MRRL) has emerged as a promising solution, enabling the learning of complex coordination behaviors while maintaining scalability.

Traditional approaches to multi-robot coordination have a rich history and offer robust solutions to standardized problems across a variety of domains~\cite{8956244,mrs_agriculture,mrs_surveillance,queralta2020collaborative}. However, these approaches often rely on explicit specification of generalizable coordination mechanisms, which either requires considerable expertise or is virtually impossible in complex unstructured domains. Learning-based techniques, such as multi-robot reinforcement learning (MRRL), are emerging as promising alternatives that can offer a generalizable recipe to enable effective multi-robot coordination across various applications (e.g., scheduling~\cite{wang2022heterogeneous}, warehouse automation~\cite{10802813}, and autonomous driving~\cite{bhalla2020deep,candela2022transferring}).

%%% Issues with MRRL, MARL, bridging function of MARBLER, No GPU parallelization
Despite its potential, learning-based algorithms are seldom deployed and evaluated on physical multi-robot platforms. Even when physical deployments are undertaken, they tend to rely on bespoke and closed platforms and testbeds, limiting reproducibility and benchmarking. 
% the evaluation of MRRL algorithms on physical robot platforms remains limited and is often performed on ad-hoc platforms. 
In stark contrast, the multi-\textit{agent} reinforcement learning (MARL) research community outside of robotics benefits from standardized and structured benchmarks, simulation, and learning environments (e.g., Multi-Agent Particle Environment (MPE)~\cite{mpe-ref}, StarCraft Multi-Agent Challenge (SMAC)~\cite{ellis2024smacv2}, and JaxMARL~\cite{jaxmarl}). However, these environments often abstract away robotics-relevant characteristics (e.g., robot dynamics and collision avoidance) and result in a considerable sim2real gap \cite{candela2022transferring}, limiting their ability to faithfully evaluate learning approaches tailored to multi-robot systems. 
% leaving a considerable sim2real gap \cite{candela2022transferring}. 

MARBLER~\cite{torbati2023marbler} is a recent framework that attempts to meet this critical need. MARBLER acts as a bridge between the standardized infrastructure designed for MARL algorithms and the Robotarium~\cite{robotarium} (an open and remotely-accessible multi-robot testbed).
% developed by Georgia Tech. 
However, MARBLER does not support parallelization and execution on GPU/TPU, resulting in prohibitively slow training speeds that hinder adoption and comprehensive evaluations. 

\begin{figure}
    \centering
    \includegraphics[width=0.9\columnwidth]{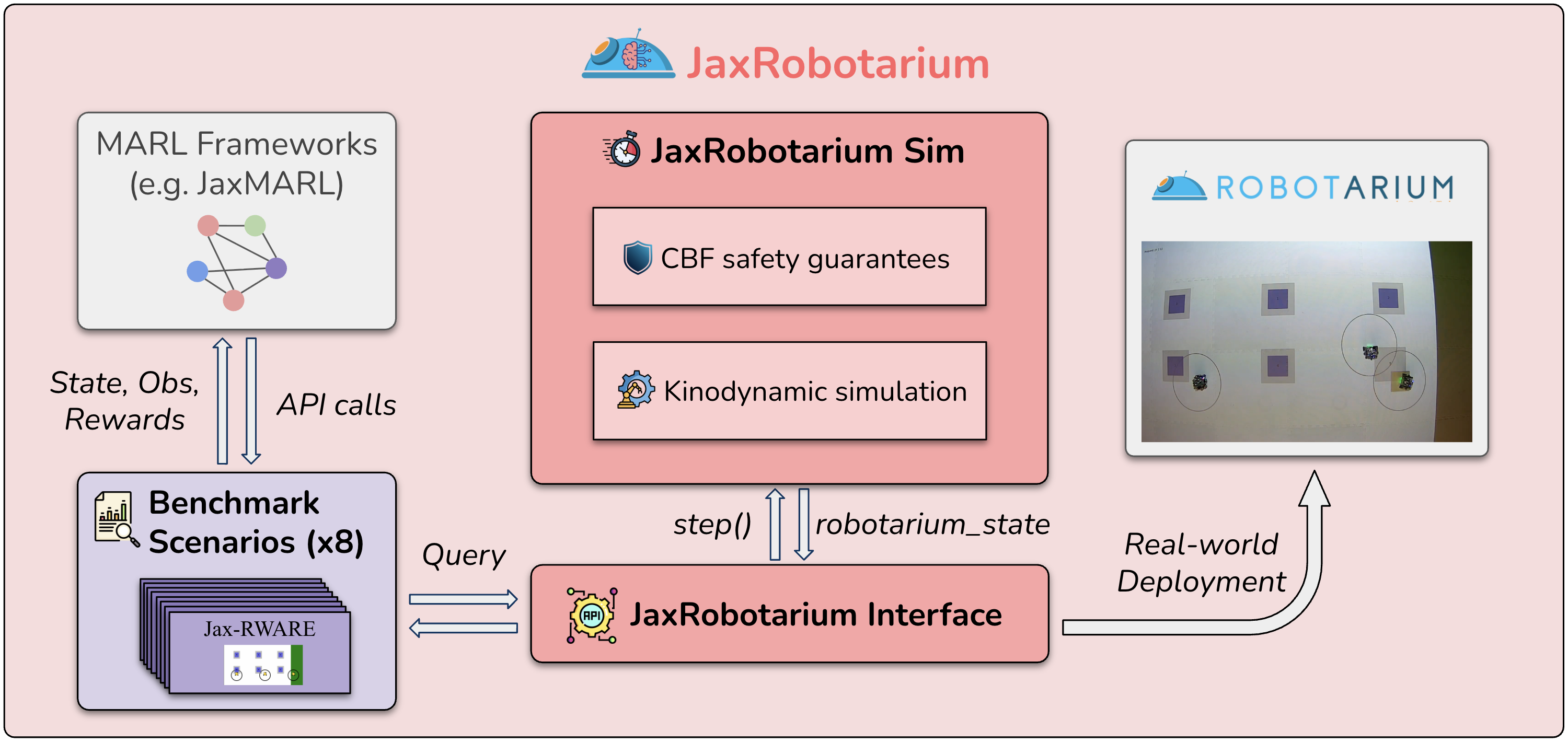}
    \caption{Overview of \frameworkName{} architecture. See \figureautorefname~\ref{fig:block_diagram} in Appendix for additional details. }
    \label{fig:overview_diagram}
    \vspace{-1em}
\end{figure}

%%% JaxRobotarium framework
\textbf{In this work, we contribute \frameworkName, an end-to-end open-source Jax-based platform designed to significantly accelerate the training and deployment of MRRL algorithms while maintaining ease-of-use and open access to hardware-based evaluations (see Fig.~\ref{fig:overview_diagram}).}
% ease of integration with the Robotarium real-robot test bed--a key merit of prior MARL-MRRL interfacing efforts. 

\frameworkName{} includes three key contributions. 
First, we contribute a new Jax-based \textit{simulator} for the Robotarium that is designed from the ground-up and modeled after the existing Robotarium Python Simulator (RPS). Our new simulator supports parallelization and GPU/TPU execution, enabling dramatic improvements in efficiency by reducing the time and computational overhead typically associated with learning multi-robot coordination in realistic settings. 
Second, we contribute a new intuitive \textit{interface} to facilitate integration with
% \frameworkName~interfaces the simulator with 
existing MARL libraries, such as JaxMARL~\cite{jaxmarl}, allowing researchers to leverage and build upon existing algorithmic implementations with minimal setup. 
Third, we contribute a new unified \textit{benchmark} for MRRL 
that currently supports 8 representative multi-robot coordination scenarios, with flexible tools to rapidly create new ones or customize the existing ones. The unique benefits of \frameworkName{} include:
\begin{flushitemize}
    \item Dramatic improvements in efficiency of training (up to 20x) and simulating (up to 150x) MRRL policies, supporting parallelization and execution on GPU/TPUs.
    \item Support for realistic simulation of robots, adhering to their dynamic constraints and safety guarantees obtained from barrier certificates.  
    \item A standardized set of multi-robot coordination scenarios implemented in Jax with real-world counterparts for rapid benchmarking.
    \item An open-source, no-cost platform that provides any user in the world with access to training and real-world deployment of MRRL algorithms. 
\end{flushitemize}

We validate the utility and benefits of \frameworkName{} through extensive efficiency evaluations, algorithm benchmarking, and comprehensive sim2real transfer experiments. We demonstrate that, compared to MARBLER~\cite{torbati2023marbler}, \frameworkName{} simulates robot trajectories up to 150 times faster and trains multi-robot policies up to 20 times faster. 
We benchmarked four MARL algorithms on the 8 scenarios provided by \frameworkName{}, revealing new insights into the algorithms' utility across different multi-robot coordination scenarios. 
We systematically studied the sim2real gap of \frameworkName{} using data from over 200 real-world deployments of multi-robot policies that were entirely trained in simulation.
We note that \frameworkName{} is not a panacea for the sim2real gap. Rather, \frameworkName{} helps study and reduce the sim2real gap by enabling rapid training and deployment. 
Our analysis reveals that while many task-algorithm combinations result in negligible performance differences when deployed on hardware, certain combinations of MARL algorithms and task characteristics exhibit noticeable performance differences. However, we show that such differences can be largely mitigated through simple randomization techniques such as noisy actions.
% , indicating that sim2real research is still useful despite the existence of \frameworkName{}.
% \skailas{In particular,} we showed that domain randomization using noisy actions aids in robustifying the policies and thus narrowing the sim2real gap.
We believe \frameworkName{} will serve as a powerful tool for the multi-robot systems community to advance the development, benchmarking, and sim2real transfer of learning algorithms.

%===============================================================================
\section{Related Work}
The platforms to facilitate learning multi-robot coordination strategies should have the following characteristics: (i) support efficient training and evaluation of policies \cite{jaxmarl, bettini2022vmas}; (ii) have realistic simulations of robots \cite{candela2022transferring, robotarium}, taking into account their dynamic constraints and the need for lower-level collision avoidance; (iii) have a convenient interface with common MARL algorithms and benchmarking tasks \cite{jaxmarl,torbati2023marbler,bettini2024benchmarl}. Centering around these key requirements, we evaluate and compare available MRRL and MARL frameworks in Table~\ref{tab:features_different_platforms}. The Multi-Agent Particle Environment (MPE)~\cite{mpe-ref} is a widely used framework for evaluating MARL algorithms on a diverse suite of 2D tasks. While MPE was enhanced by successors including VMAS~\cite{bettini2022vmas}, BenchMARL~\cite{bettini2024benchmarl}, and JaxMARL~\cite{jaxmarl}, none of them realistically reflects the dynamics of real robots, impeding straightforward sim2real evaluation of MRRL policies.
IsaacGym \cite{makoviychuk2021isaac} and IsaacLab \cite{mittal2023orbit} offer high-fidelity simulation but are geared toward single-robot learning across varied embodiments. Multi-robot learning in these frameworks requires additional infrastructure for integration with MARL libraries, collision avoidance, benchmark tasks, and a sim2real pipeline, none of which are natively provided.
Thus, to better evaluate MRRL policies, researchers must build task-specific test platforms \cite{chen2023multirobolearn, srinivasa2019mushr, 7989179} or reproduce previously proposed test beds, such as Cambridge RoboMaster~\cite{blumenkamp2024cambridge}. As a contrastive line of efforts, the Robotarium~\cite{robotarium} is a publicly available, multi-robot testbed that supports testing the coordination of up to 20 GRITSBots~\cite{GRITSBots} robots. The Robotarium provides a simulator with implemented collision avoidance using control barrier functions~\cite{cbf-mrs-li-wang} and offers sim2real evaluation through its hardware test bed. It is the only publicly available and free-to-use multi-robot hardware testbed in the world, and has been widely used for multi-robot research in areas including task allocation \cite{notomista}, heterogeneous coordination \cite{seraj2024heterogeneous}, motion planning \cite{sun2022multi}, and safety \cite{chen2020guaranteed}. To bridge the gap between MARL and a real multi-robot testbed,~\cite{torbati2023marbler} contributes MARBLER, which is an open platform that promotes the comprehensive evaluation of MRRL algorithms on the Robotarium. Unfortunately, MARBLER only runs on the CPU without parallelization, resulting in prolonged training and tedious development cycles. \frameworkName{} builds upon the philosophy of MARBLER, marrying it with the computational efficiency offered by Jax and GPU parallelization.   

\begin{table*}[h]
    \centering
    \resizebox{\textwidth}{!}{%
    \begin{tabular}{l|c|c|c|c|c|c}
        \textbf{Platform} & \textbf{Train On} & \textbf{Robot} & \textbf{Safety}  & \textbf{Sim2Real} & \textbf{MARL} & \textbf{Open Access} \\
        & \textbf{GPU/TPU} & \textbf{Dynamics} & \textbf{Guarantee} & \textbf{Capabilities} & \textbf{Support} & \textbf{Hardware} \\
        \hline 
        % MPE~\cite{mpe-ref} & \cmark & \xmark & \xmark & \xmark & \cmark \\
        % \hline 
        % VMAS~\cite{bettini2022vmas} & \cmark & \xmark & \xmark & \xmark & \cmark \\
        % \hline
        % SMAC~\cite{ellis2024smacv2} & \cmark & \xmark & \xmark & \xmark & \cmark \\ 
        % \hline
        MARL Platforms (MPE~\cite{mpe-ref}, SMAC~\cite{ellis2024smacv2},
        VMAS~\cite{bettini2022vmas},
         & \cmark & \xmark & \xmark & \xmark & \cmark & N/A \\ 
        % , & & & & & \\ 
         BenchMARL~\cite{bettini2024benchmarl}, JaxMARL~\cite{jaxmarl}, Google Research Football~\cite{kurach2020google}) & & & & & & \\ 
        \hline
        % JaxMARL~\cite{jaxmarl} & \cmark & \xmark & \xmark & \xmark & \cmark  \\
        % \hline
        % Google Research Football~\cite{kurach2020google} & \cmark & \xmark & \xmark & \xmark & \cmark \\
        % \hline
        IsaacGym \cite{makoviychuk2021isaac}, IsaacLab \cite{mittal2023orbit} & \cmark & \cmark & \xmark & \cmark & \xmark & \xmark \\
        \hline
        Cambridge RoboMaster \cite{blumenkamp2024cambridge} & \cmark & \cmark & \xmark & \cmark & \cmark & \xmark \\
        \hline
        MuSHR \cite{srinivasa2019mushr} & \cmark & \cmark & \cmark & \cmark & \xmark & \xmark \\ 
        \hline
        Duckietown \cite{7989179,candela2022transferring} & \cmark & \cmark & \xmark & \cmark & \cmark & \xmark
        \\ \hline
        Robotarium~\cite{robotarium} & \xmark & \cmark & \cmark & \cmark & \xmark & \cmark \\ 
        \hline 
        MARBLER~\cite{torbati2023marbler} & \xmark & \cmark & \cmark & \cmark & \cmark & \cmark \\
        \hline
        \textbf{\frameworkName{} (ours)} & \cmark & \cmark & \cmark & \cmark & \cmark & \cmark \\
    \hline
    \end{tabular}}
    \caption{\frameworkName{} versus existing frameworks for multi-robot/multi-agent learning. }
    \label{tab:features_different_platforms}
    \vspace{-1em}
\end{table*}

\section{\frameworkName{} Platform}
% Multi-agent reinforcement learning (MARL)  has facilitated learning multi-agent coordination policies that succeed in complex domains, such as StarCraft \cite{ellis2024smacv2}. This success has resulted in the multi-robot community adopting MARL algorithms to learn Multi-Robot coordination in tasks such as social navigation \cite{escudie2024attention}, material transport \cite{howell2024generalization}, and sensor networks \cite{howell2024generalization}. However, sim2real evaluation represents a consistent challenge in evaluating MARL algorithms for multi-robot platforms, requiring researchers to build task-specific test platforms \cite{chen2023multirobolearn} or reproduce previously proposed test beds, such as Cambridge RoboMaster \cite{blumenkamp2024cambridge}. The Multi-Agent RL Benchmark and Learning Environment
% for the Robotarium (MARBLER) aims to simplify sim2real evaluation by providing a pipeline for training and deploying multi-robot policies by bridging existing learning libraries with the existing Robotarium Python Simulator (\texttt{RPS}). Unfortunately, its usefulness is severely limited due to not supporting GPU execution or parallelization, which are key components in the recent success and speed-up of Multi-Agent RL.
\frameworkName{} is designed to achieve the training speeds of Multi-Agent RL in Multi-Robot RL while offering standardized, public, and free sim2real evaluation through the Robotarium. To this end, \frameworkName{} consists of the following key components,
\begin{flushenumerate}
    \item A Jax-based simulator for the Robotarium, compatible with parallelization and GPU/TPU execution.
    \item A platform for Multi-Robot RL, which we integrated with JaxMARL \cite{jaxmarl} to enable training with several SOTA MARL algorithms.
    \item An implementation of all four MARBLER tasks, and four newly implemented tasks.
    \item A direct deployment pipeline for sim2real evaluation on the Robotarium.
\end{flushenumerate}
% We dedicate the remainder of this paper to discussing and evaluating each component. 
We provide a full overview of the architecture in the Appendix (Figure \ref{fig:block_diagram}).
% demonstrate that these components provide up to a 20x speedup over MARBLER (Section .

% \begin{figure}
%     \centering
%     \includegraphics[width=\columnwidth]{figures/jax-robotarium-arch.png}
%     \caption{\frameworkName{} architecture, colored components are novel.}
%     \label{fig:block_diagram}
% \end{figure}

%===============================================================================
\section{Jax-based Robotarium Simulator}
\subsection{Design}
We contribute \texttt{Jax-RPS}, a Jax simulator for the Robotarium. \texttt{Jax-RPS} uses a unicycle dynamics model to faithfully simulate the behavior of the Robotarium's GRITSBots \cite{pickem2015gritsbot}. To control the GRITSBots, \texttt{Jax-RPS} implements single integrator and unicycle controllers to drive robots to desired positions/poses. To enforce the safety constraints often required in real-world deployment, \texttt{Jax-RPS} implements control barrier certificates for collision avoidance, which solve the quadratic program described in \cite{robust-cbf}. Finally, all functionality in \texttt{Jax-RPS} is designed to be compatible with Jax's \texttt{jit} compilation for accelerated execution and \texttt{vmap} for parallelization. Overall, this results in \texttt{Jax-RPS} being optimized for data-driven paradigms, such as Multi-Agent RL, while maintaining its faithfulness to real-world conditions and deployment considerations.

\subsection{Evaluation}
\begin{figure*}[t]
    \centering
    % First: Table
    \begin{minipage}[b]{0.60\textwidth}
        \centering
        \resizebox{\textwidth}{!}{%
        \begin{tabular}{l|l|cc}
            \textbf{Controller} & \textbf{Simulator} & \textbf{Wall Time} & \textbf{Trajectory} \\
            & &\textbf{(ms $\downarrow$)} &\textbf{Error ($\downarrow$)} \\
            \hline
            Position & Default & 1425.28 $\pm$ 4.79 & N/A \\
                & JaxRobotarium & \textbf{9.18 $\pm$ 0.10} & 0.00 $\pm$ 0.00 \\
            \hline
            Pose & Default & 1713.87 $\pm$ 5.39 & N/A \\
                & JaxRobotarium & \textbf{11.42 $\pm$ 0.07} & 0.00 $\pm$ 0.00 \\
            \hline
        \end{tabular}}
        \captionof{table}{\texttt{Jax-RPS} simulates the same trajectories significantly faster than \texttt{RPS}. Trajectory errors are computed for \texttt{Jax-RPS} against the trajectory for the identical scenario initial conditions simulated in \texttt{RPS}.}
        \label{tab:sim-mismatch-results}
    \end{minipage}
    \hfill
    % Second: Figure
    \begin{minipage}[b]{0.36\textwidth}
        \centering
        \includegraphics[width=0.8\textwidth]{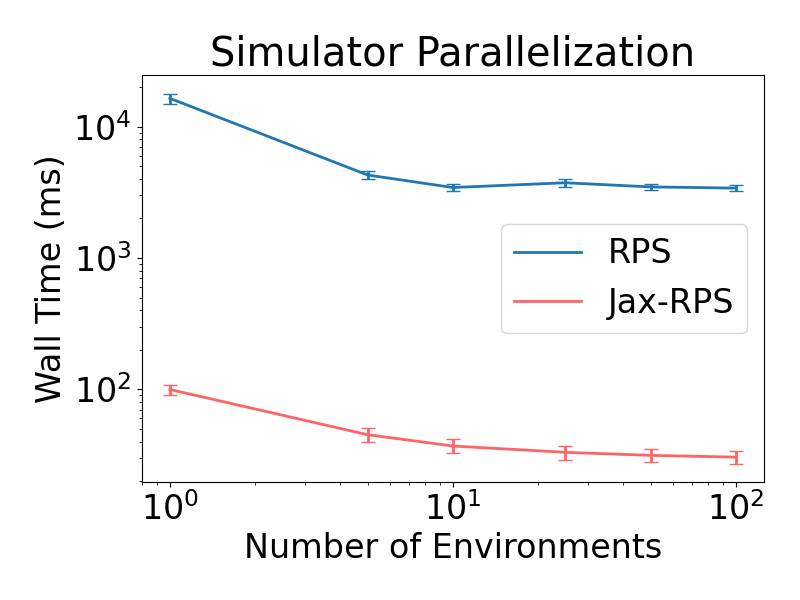} % or .png etc.
        \caption{\texttt{Jax-RPS} maintains 100x or more speed up over \texttt{RPS} across increasing degrees of parallelization.}
        \label{fig:sim-parallel-figure}
    \end{minipage}
\end{figure*}

To validate the faithfulness and optimization of \texttt{Jax-RPS}, we construct the Random Waypoint Navigation scenario where four robots must each navigate to a randomly selected waypoint before being assigned a new one. We conduct the following experiments:
\begin{flushenumerate}
    \item We measure faithfulness by simulating Random Waypoint Navigation for 10K timesteps and computing the difference in robot trajectories between \texttt{Jax-RPS} and \texttt{RPS} across unicycle position and pose controllers. We report mean wall time and trajectory mismatch with associated standard error averaged across 30 trials in \tableautorefname~\ref{tab:sim-mismatch-results}.
    \item We measure optimization performance by simulating Random Waypoint Navigation for 100K timesteps, where the timesteps are collected across multiple environments in parallel. We modified \texttt{RPS} to support multi-processing for parallelization. We report mean wall time across the number of environments with associated standard deviation averaged over 30 trials in \figureautorefname~\ref{fig:sim-parallel-figure}.
\end{flushenumerate}
In all trials, we randomize the order of the controller-simulator conditions to avoid ordering biases. The above experiments were run on an M4 Pro CPU and 24 GB of RAM. 
As seen in Table \ref{tab:sim-mismatch-results}, \texttt{Jax-RPS} simulates trajectories that are identical to \texttt{RPS} with a 140- 150x speedup over \texttt{RPS} when run on CPU. Furthermore, in Figure \ref{fig:sim-parallel-figure} we observe \texttt{Jax-RPS} continues to benefit from increased parallelization, with its wall time to collect 100K timesteps monotonically decreasing as the number of environments increases. In contrast, \texttt{RPS} suffers slightly at higher degrees of parallelization, where the overhead of managing the multiple processes outweighs the benefit of executing the environments in parallel.

% \begin{table*}[h]
%     \centering
%     \begin{tabular}{c|c|c|c}
%         \textbf{Environments} & \textbf{Simulator} & \textbf{Wall Time (ms $\downarrow$)} & \textbf{Step Time (ms $\downarrow$)} \\
%         \hline
%         1 & Python & 4025.721 $\pm$ 1632.511 & 0.403 $\pm$ 0.163 \\
%             & Jax & \textbf{32.783 $\pm$ 22.909} & \textbf{0.003 $\pm$ 0.002} \\
%         \hline
%         5 & Python & 1689.913 $\pm$ 346.188 & 0.169 $\pm$ 0.035 \\
%             & Jax & \textbf{17.355 $\pm$ 12.700} & \textbf{0.002 $\pm$ 0.001} \\
%         \hline
%         10 & Python & 1758.737 $\pm$ 215.992 & 0.176 $\pm$ 0.022 \\
%             & Jax & \textbf{11.527 $\pm$ 7.829} & \textbf{0.001 $\pm$ 0.001} \\
%         \hline
%     \end{tabular}
%     \caption{Simulation Parallelization Results}
%     \label{tab:sim-parallel-results}
% \end{table*}

%===============================================================================
\section{Benchmark Scenarios}
We provide a diverse set of tasks to train on within \frameworkName. First, we re-implement the original MARBLER tasks in our framework, as described below.
\begin{flushitemize}
    \item \textit{Arctic Transport}: Two drone robots that observe their surrounding terrain must help guide two traversal robots, one faster on ice and slower on water, and vice versa. The team is successful if both traversal robots reach the goal zone.
    \item \textit{Predator Capture Prey $\rightarrow$ Discovery}: Sensing robots that sense landmarks within the environment collaborate with tagging robots that tag landmarks within the environment. Task performance is measured by the number of landmarks tagged. We note that this task has been renamed to Discovery for clarity.
    \item \textit{Material Transport}: Robots with varying speeds and capacities must collaborate to unload two material depots, one further with less material, and one closer with more material. Task performance is measured by the amount of material remaining to be unloaded.
    \item \textit{Warehouse}: Robots assigned to a zone type must collectively maximize the number of deliveries between their assigned zone type. Task performance is measured by the number of deliveries
\end{flushitemize}
In addition to the re-implementations above, we designed and implemented four new tasks:
% , as described below.
\begin{flushitemize}
    \item \textit{Navigation (MAPF)}: Robots must navigate to independent goals within the environment. The team is successful if all robots are on their goals by the end of the rollout.
    \item \textit{Foraging}: Inspired by grid world Level-Based Foraging \cite{lbf-ref}, robots with varying levels of foraging ability must collaborate to forage all resources in the environment. Task performance is measured by the number of resources foraged.
    \item \textit{Continuous-RWARE}: Inspired by the grid world RWARE \cite{rware-ref}, robots must collaborate to fulfill dropoffs of the requested shelves. Task performance is measured by the number of requested shelves dropped off.
    \item \textit{Predator-Prey}: Robots must collaborate to tag a more agile prey, where the prey is controlled by the heuristic proposed in \cite{peng2021facmac}. Task performance is measured by the number of tags on the prey.
\end{flushitemize}
Visualizations of all scenarios can be seen in Figure \ref{fig:sim2real}, and detailed descriptions are in Appendix \ref{appendix:scenario-details}.

%===============================================================================
\section{Jax-Robotarium Interface}
\subsection{Design}
To enable the training and deployment of multi-robot policies with our simulator, we build a learning interface consisting of 5 main components:
\begin{flushitemize}
    \item \texttt{RobotariumEnv} is the main interface structuring input to and output from \texttt{Jax-RPS}. It maintains a generalizable representation of the environment state for the Robotarium and provides visualization utilities to construct a consistent base visualization across all tasks.
    \item \texttt{HetMananger} provides an interface for injecting heterogeneity into multi-robot environments. It supports common heterogeneous representations: robot IDs, class IDs, capability vectors, and can be extended to include more. It manages the robot's awareness of heterogeneity, where the observation space can be configured to include any representation.
    \item \texttt{ControllerManager} handles translating policy actions into commands to the robots. If a policy is designed to output waypoints, these waypoints are processed by the configured controller into unicycle velocity commands. If barrier functions are enabled, the unicycle velocity commands are then modified to guarantee collision avoidance. 
    \item \texttt{Scenario} interface is a standard structure for defining new tasks within \frameworkName{}, aligned with the structures used in other learning frameworks such as VMAS \cite{bettini2022vmas}, JaxMARL \cite{jaxmarl}, and MARBLER \cite{torbati2023marbler}. 
    % At its core, it provides the interface for defining a task-specific observation, reward, and transition function.
    \item \texttt{Deploy} is a module that streamlines deployment to the Robotarium by compiling all the necessary files for an experiment into a folder, which can then be directly uploaded and run on the Robotarium test bed. Since the Robotarium test bed does not support Jax, the deployment module performs model conversion and file sanitation to ensure compatibility.
\end{flushitemize}
We choose JaxMARL as the MARL framework, given its SOTA performance and extensive baseline algorithm implementations. However, note that any Jax multi-agent learning framework designed for the output structure of the \texttt{Scenario} interface is compatible with \frameworkName{}.

\subsection{Evaluation}
To validate the utility of our framework, we compare the training performance and efficiency of \frameworkName{} with no parallelization (1 env) and parallelization (8 envs) against MARBLER on the four MARBLER scenarios: Arctic Transport, Discovery, Material Transport, and Warehouse. We report the episodic returns and timesteps simulated against the wall time of the fastest run, averaged across 3 seeds. All experiments are run on an Intel i7-12700KF and NVIDIA RTX 3070.

\begin{figure}
    \centering
    \begin{subfigure}{0.24\columnwidth}
      \includegraphics[width=\columnwidth]{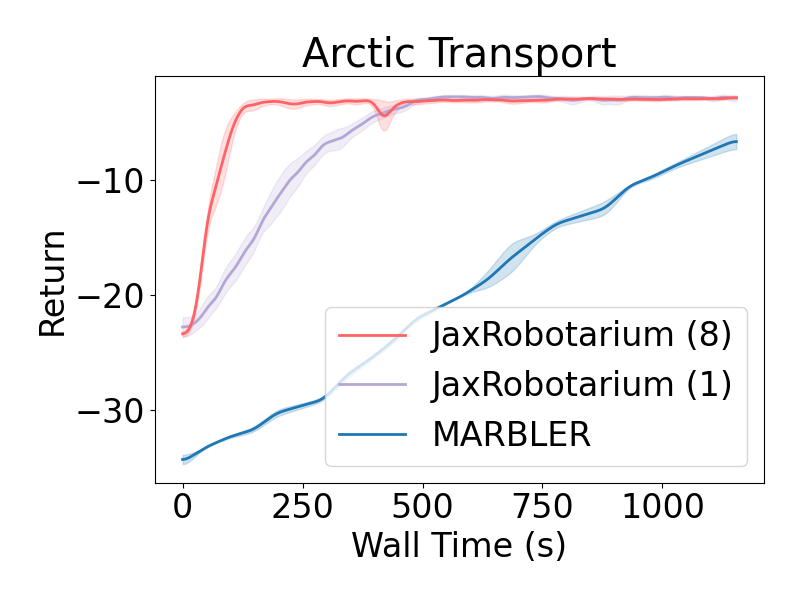}
    \end{subfigure}
    \begin{subfigure}{0.24\columnwidth}
      \includegraphics[width=\columnwidth]{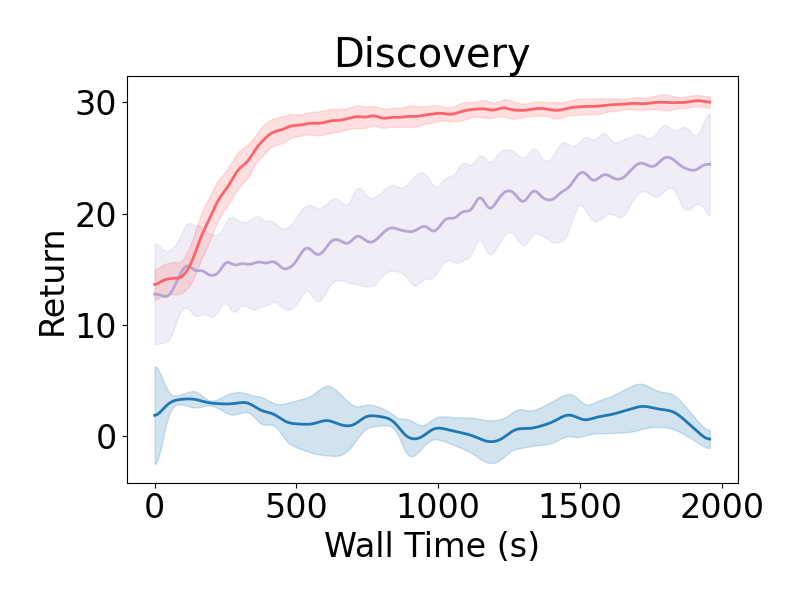}
    \end{subfigure}
    \begin{subfigure}{0.24\columnwidth}
      \includegraphics[width=\columnwidth]{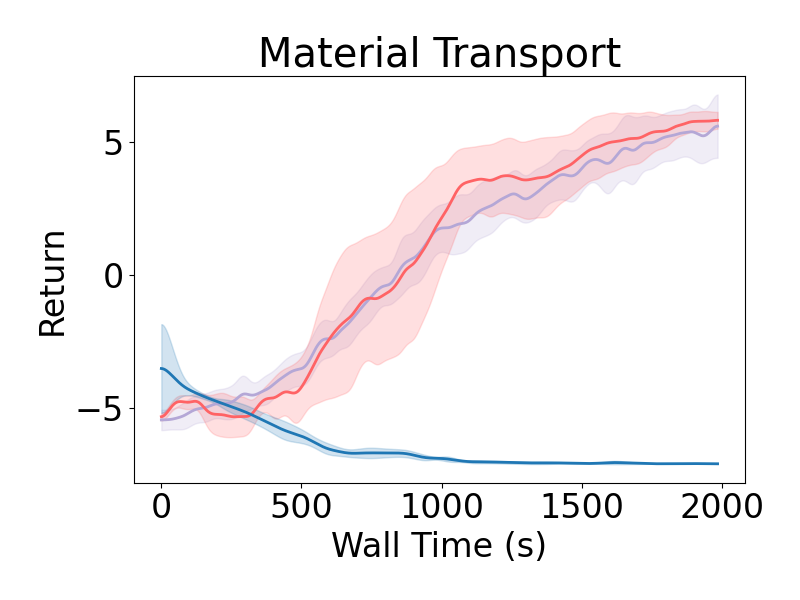}
    \end{subfigure}
    \begin{subfigure}{0.24\columnwidth}
      \includegraphics[width=\columnwidth]{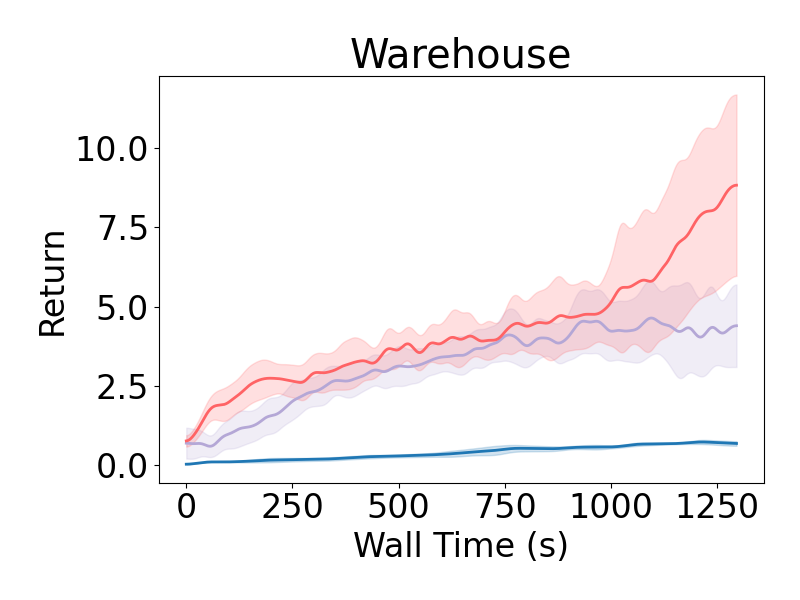}
    \end{subfigure}

    \begin{subfigure}{0.24\columnwidth}
      \includegraphics[width=\columnwidth]{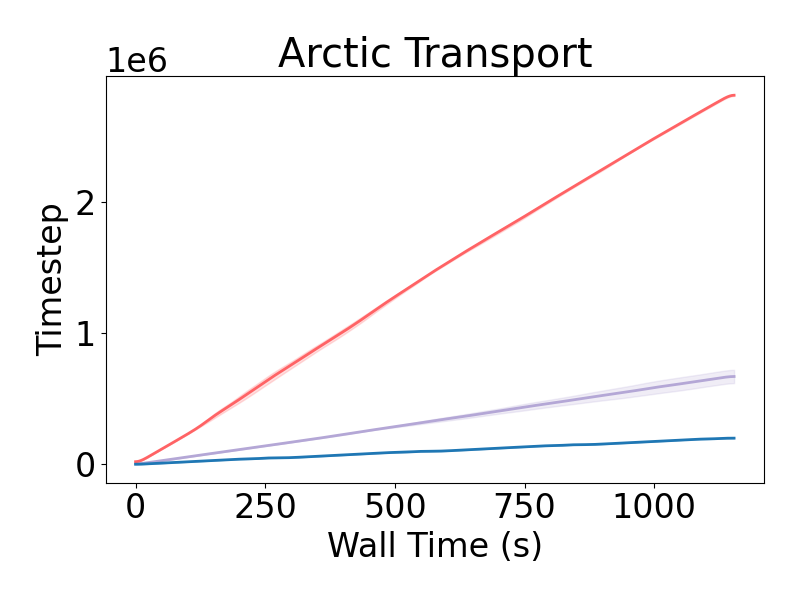}
    \end{subfigure}
    \begin{subfigure}{0.24\columnwidth}
      \includegraphics[width=\columnwidth]{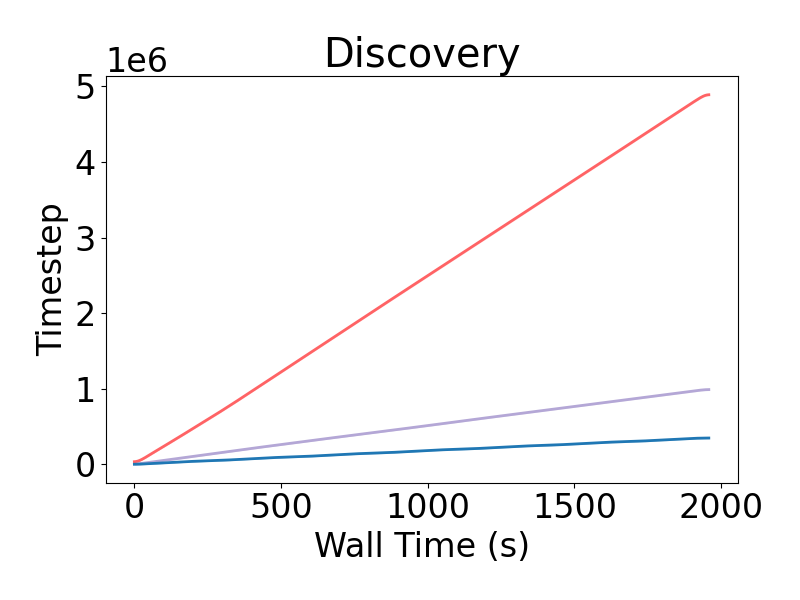}
    \end{subfigure}
    \begin{subfigure}{0.24\columnwidth}
      \includegraphics[width=\columnwidth]{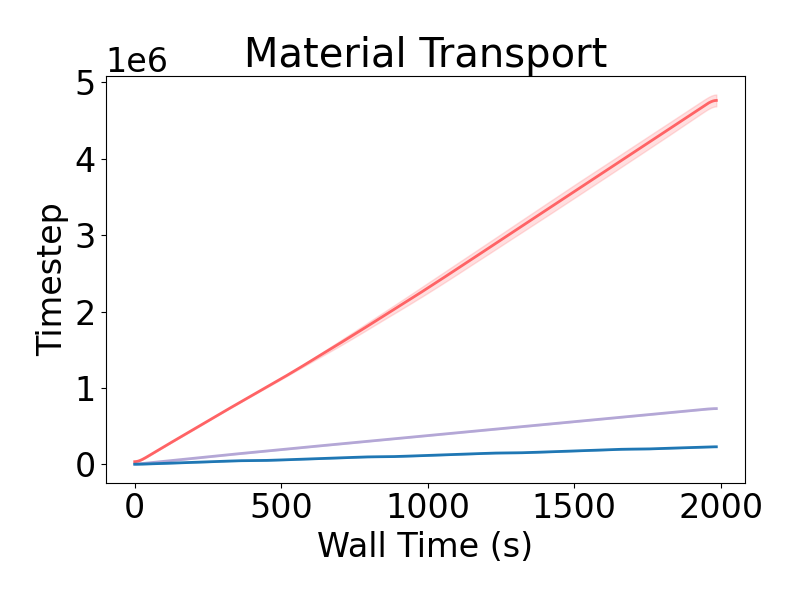}
    \end{subfigure}
    \begin{subfigure}{0.24\columnwidth}
      \includegraphics[width=\columnwidth]{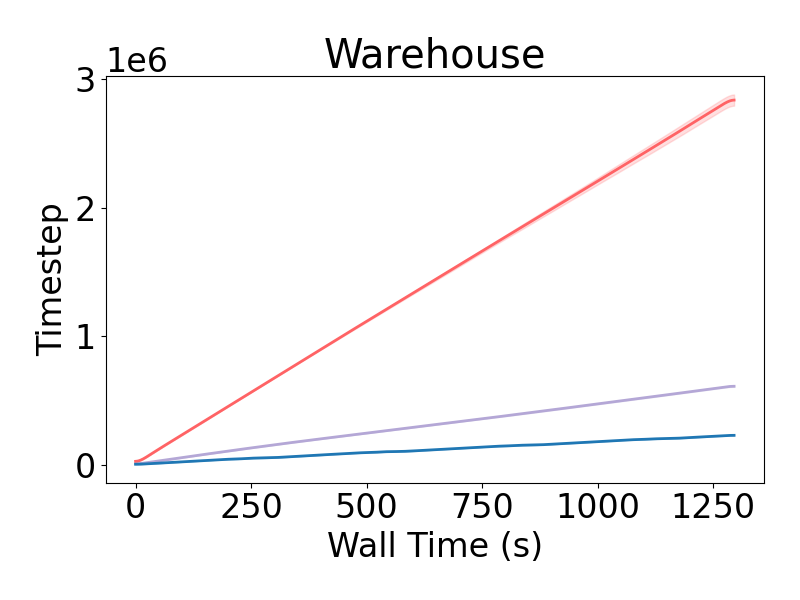}
    \end{subfigure}
    
    \caption{(Top) We plot the mean returns achieved in both platforms against the minimum training wall time. (Bottom) We plot the time taken to simulate timesteps across both platforms. We find that \frameworkName{} is significantly more efficient in training multi-robot policies than MARBLER. Results depict smoothed mean (solid line) with standard deviation (shaded).}
    \label{fig:exp-train-efficiency}
    \vspace{-1em}
\end{figure}

% \subsection{Results}
In Figure \ref{fig:exp-train-efficiency}, we find that \frameworkName{} can train for up to 20x more timesteps and achieve consistently higher returns compared to MARBLER given the same amount of time. As expected, the support for GPU execution, parallelization, and just-in-time compilation significantly accelerates training across all tasks, bridging the gap between the Robotarium and modern MARL algorithms.

% \begin{enumerate}
%     \item \textbf{Setup}: Train MARBLER and JaxRobotarium using QMIX on all originally implemented MARBLER tasks.
%     \item \textbf{Independent Variables}: \begin{flushitemize}
%         \item Platform: JaxRobotarium, MARBLER
%         \item Number of parallel environments: 1 (MARBLER), 8 (JaxRobotarium)
%         \item Number of parallel seeds: 1
%     \end{flushitemize}
%     \item \textbf{Metrics}: Training Returns, Wall Time
%     \item \textbf{Presentation}: To report wall time results, a figure resembling Figure \ref{fig:training-wall-time}. To report training return results, graph single seed training returns with the x-axis being wall time, across tasks.
%     \item \textbf{Procedure}: \begin{enumerate}
%         \item For JaxRobotarium training runs, ablate widths of 128, 256, and 512 on single seed. Use the best-performing width to run final training runs. Log metrics in wandb.
%         \item For MARBLER training runs, adjust the task to reflect implementation in JaxRobotarium. Use the previous architecture parameters to run experiments. Log metrics in wandb. 
%     \end{enumerate}
% \end{enumerate}

%===============================================================================
\section{Benchmarking Learning Algorithms}
% \subsection{Design}

% Thus, deploying an experiment is as simple as supplying the module with the configuration file corresponding to the experiment to be run on the Robotarium.

\subsection{Experiment setup}
We demonstrate the benchmarking and sim2real evaluation capabilities of \frameworkName{} by training 5 policies in parallel for all tasks using the following SOTA MARL approaches:
\begin{flushitemize}
    \item PQN \cite{pqn}: a recent multi-agent Q-Learning approach designed to perform Q-updates efficiently by collecting non-correlated transitions in parallel across vectorized environments (we use 256), eliminating the need for a replay buffer.
    \item QMIX \cite{qmix}: an established multi-agent Q-learning algorithm where decentralized Q-networks predict agent Q-values aggregated into a global Q-value by a hypernetwork-generated centralized critic conditioned on the global state.
    \item MAPPO \cite{mappo-ippo}: an established multi-agent policy-gradient algorithm that learns decentralized action policies, regularized by a centralized critic conditioned on the global state.
    \item IPPO \cite{mappo-ippo}: a multi-agent policy-gradient algorithm that learns decentralized action policies. Unlike MAPPO, IPPO uses a decentralized critic conditioned on the agent's local observation.
\end{flushitemize}
See Appendix \ref{appendix:training-details} for further training details. We report mean episodic reward, collisions, and a task-performance metric with associated standard errors. Metrics are averaged across the 5 trained policies for 100 episodes each. For real-world experiments, we deploy the best-performing policy on the Robotarium hardware testbed and report metrics averaged across 5 unique instantiations, for a total of 160 experiments.
% \begin{enumerate}
%     \item \textbf{Setup}: Train and deploy policies for all \frameworkName{} tasks.
%     \item \textbf{Independent Variables}: \begin{flushitemize}
%         \item Learning Algorithm: QMIX, PQN, MAPPO, IPPO
%         \item Deployment: Simulation, Robotarium (3 episodes per condition?)
%     \end{flushitemize}
%     \item \textbf{Metrics}: Training Returns, Task specific metrics
%     \item \textbf{Presentation}: Report using a table copying the structure of Table II in \cite{torbati2023marbler}
% \end{enumerate}
\subsubsection{Results}

\begin{wraptable}{r}{0.7\textwidth}
\vspace{-1em}
\resizebox{0.7\textwidth}{!}{%
\begin{tabular}{ll|cccc}

\textbf{Scenario} & \textbf{Metric} & \multicolumn{4}{c}{\textbf{Training-Seed Benchmark}} \\
 & & \textbf{PQN} & \textbf{QMIX} & \textbf{MAPPO} & \textbf{IPPO} \\
\hline

\textbf{Arctic} & Reward & -3.6 $\pm$ 0.04 & -3.4 $\pm$ 0.02 & -3.5 $\pm$ 0.04 & -3.5 $\pm$ 0.04 \\
\textbf{Transport} & Success Rate & 1.0 $\pm$ 0.00 & 1.0 $\pm$ 0.00 & 1.0 $\pm$ 0.00 & 1.0 $\pm$ 0.00 \\
 & Collisions & 0.0 $\pm$ 0.00 & 0.0 $\pm$ 0.00 & 0.0 $\pm$ 0.00 & 0.0 $\pm$ 0.00 \\
\hline
\textbf{Discovery} & Reward & 26.6 $\pm$ 0.20 & 29.8 $\pm$ 0.14 & 24.1 $\pm$ 0.32 & 30.0 $\pm$ 0.15 \\
 & Landmarks Tagged & 5.1 $\pm$ 0.04 & 5.6 $\pm$ 0.03 & 4.5 $\pm$ 0.06 & 5.7 $\pm$ 0.03 \\
 & Collisions & 0.0 $\pm$ 0.00 & 0.0 $\pm$ 0.00 & 0.0 $\pm$ 0.00 & 0.0 $\pm$ 0.00 \\
\hline
\textbf{Material} & Reward & -5.0 $\pm$ 0.06 & 4.6 $\pm$ 0.07 & 1.3 $\pm$ 0.08 & 1.3 $\pm$ 0.09 \\
\textbf{Transport} & Material Left & 65.1 $\pm$ 0.75 & 4.2 $\pm$ 0.31 & 15.7 $\pm$ 0.72 & 17.1 $\pm$ 0.71 \\
 & Collisions & 0.0 $\pm$ 0.00 & 0.0 $\pm$ 0.00 & 0.0 $\pm$ 0.00 & 0.0 $\pm$ 0.00 \\
\hline
\textbf{Warehouse} & Reward & 8.4 $\pm$ 0.20 & 6.1 $\pm$ 0.16 & 3.9 $\pm$ 0.02 & 8.0 $\pm$ 0.17 \\
 & Deliveries & 1.7 $\pm$ 0.06 & 0.9 $\pm$ 0.05 & 0.0 $\pm$ 0.00 & 1.7 $\pm$ 0.05 \\
 & Collisions & 0.0 $\pm$ 0.00 & 0.0 $\pm$ 0.00 & 0.0 $\pm$ 0.00 & 0.0 $\pm$ 0.00 \\
\hline
\textbf{Navigation} & Reward & -18.3 $\pm$ 0.39 & -17.0 $\pm$ 0.17 & -13.7 $\pm$ 0.00 & -12.7 $\pm$ 0.16 \\
\textbf{(MAPF)} & Success Rate & 0.6 $\pm$ 0.39 & 0.8 $\pm$ 0.17 & 1.0 $\pm$ 0.00 & 1.0 $\pm$ 0.16 \\
 & Collisions & 0.0 $\pm$ 0.00 & 0.0 $\pm$ 0.00 & 0.0 $\pm$ 0.00 & 0.0 $\pm$ 0.00 \\
\hline
\textbf{Foraging} & Reward & 1.8 $\pm$ 0.12 & 4.5 $\pm$ 0.13 & 0.3 $\pm$ 0.06 & 7.1 $\pm$ 0.21 \\
 & Resources Foraged & 0.3 $\pm$ 0.02 & 0.8 $\pm$ 0.03 & 0.1 $\pm$ 0.01 & 1.3 $\pm$ 0.01 \\
 & Collisions & 0.0 $\pm$ 0.00 & 0.0 $\pm$ 0.00 & 0.0 $\pm$ 0.00 & 0.0 $\pm$ 0.00 \\
\hline
\textbf{Predator} & Reward & 58.8 $\pm$ 2.45 & 6.3 $\pm$ 0.65 & 11.0 $\pm$ 1.49 & 46.9 $\pm$ 3.15 \\
\textbf{Prey} & Tags & 5.9 $\pm$ 0.25 & 0.6 $\pm$ 0.06 & 1.1 $\pm$ 0.15 & 4.7 $\pm$ 0.31 \\
 & Collisions & 0.0 $\pm$ 0.00 & 0.0 $\pm$ 0.00 & 0.0 $\pm$ 0.00 & 0.0 $\pm$ 0.00 \\
\hline
\textbf{Continuous} & Reward & 5.9 $\pm$ 0.78 & 3.6 $\pm$ 0.66 & 0.0 $\pm$ 0.59 & 1.1 $\pm$ 0.53 \\
\textbf{RWARE} & Shelf Dropoffs & 5.9 $\pm$ 0.02 & 3.6 $\pm$ 0.02 & 0.0 $\pm$ 0.00 & 1.1 $\pm$ 0.01 \\
 & Collisions & 0.0 $\pm$ 0.00 & 0.0 $\pm$ 0.00 & 0.0 $\pm$ 0.00 & 0.0 $\pm$ 0.00 \\

\hline
\end{tabular}%
}
\caption{Performance metrics across all training seeds in simulation.}
\vspace{-1em}
\label{tab:training-seed-benchmark}
\end{wraptable}

\textbf{Learning}: 
In general, we do not find any single algorithm to consistently outperform all others across our 8 scenarios.
Consistent with recent findings~\cite{pqn}, we find that Q-Learning with PQN performs comparably or even outperforms Q-Learning with QMIX, with Material Transport being a notable exception (see Table \ref{tab:training-seed-benchmark}). Given that PQN achieves this performance without maintaining a replay buffer, these results support PQN as a promising alternative to established Q-Learning methods for multi-robot settings, in which maintaining a large replay buffer could be intractable due to high-dimensional state spaces and combinatorial space complexity.
% where the memory requirement of a replay buffer can be prohibitive for large teams with high-dimensional state spaces in realistic environments.

Further, we note that independent learning (IPPO) is surprisingly stronger in our multi-robot coordination scenarios compared to its centralized counterpart (MAPPO). We believe this is likely due to the limited partial observability in the configured benchmark scenarios (e.g., robots can generally observe their neighbors except in Navigation). 
% Additionally, the burden of collision avoidance is placed on using barrier certificates. 
This suggests that in settings without significant partial observability,
% and externally enforced safety guarantees, 
independent learning is a simpler and effective alternative to centralized critic approaches, in line with prior results comparing MAPPO and IPPO on StarCraft tasks \cite{mappo-ippo}.

\begin{table*}[t]
\centering
\resizebox{\textwidth}{!}{%
\begin{tabular}{ll|cccc|cccc}
\textbf{Scenario} & \textbf{Metric} & \multicolumn{4}{c|}{\textbf{Simulated Experiments}} & \multicolumn{4}{c}{\textbf{Real-World Experiments}} \\
 & & \textbf{PQN} & \textbf{QMIX} & \textbf{MAPPO} & \textbf{IPPO} & \textbf{PQN} & \textbf{QMIX} & \textbf{MAPPO} & \textbf{IPPO} \\
\hline
\textbf{Arctic} & Reward & -3.5 $\pm$ 0.22 & -3.5 $\pm$ 0.18 & -3.7 $\pm$ 0.22 & -3.4 $\pm$ 0.18 & -2.6 $\pm$ 0.13 & -3.0 $\pm$ 0.13 & -3.8 $\pm$ 1.16 & -2.5 $\pm$ 0.09 \\
\textbf{Transport} & Success Rate & 1.0 $\pm$ 0.00 & 1.0 $\pm$ 0.00 & 1.0 $\pm$ 0.00 & 1.0 $\pm$ 0.00 & 1.0 $\pm$ 0.00 & 1.0 $\pm$ 0.00 & 0.8 $\pm$ 0.18 & 1.0 $\pm$ 0.00 \\
 & Collisions & 0.0 $\pm$ 0.00 & 0.0 $\pm$ 0.00 & 0.0 $\pm$ 0.00 & 0.0 $\pm$ 0.00 & 0.0 $\pm$ 0.00 & 0.0 $\pm$ 0.00 & 0.0 $\pm$ 0.00 & 0.0 $\pm$ 0.00 \\
\hline
\textbf{Discovery} & Reward & 26.1 $\pm$ 1.88 & 25.2 $\pm$ 1.16 & 23.6 $\pm$ 2.68 & 29.3 $\pm$ 1.25 & 24.7 $\pm$ 1.07 & 26.1 $\pm$ 1.48 & 23.9 $\pm$ 2.55 & 29.3 $\pm$ 1.25 \\
 & Landmarks Tagged & 5.0 $\pm$ 0.40 & 4.8 $\pm$ 0.18 & 4.6 $\pm$ 0.54 & 5.6 $\pm$ 0.22 & 4.6 $\pm$ 0.22 & 5.0 $\pm$ 0.27 & 4.6 $\pm$ 0.45 & 5.6 $\pm$ 0.22 \\
 & Collisions & 0.0 $\pm$ 0.00 & 0.0 $\pm$ 0.00 & 0.0 $\pm$ 0.00 & 0.0 $\pm$ 0.00 & 0.0 $\pm$ 0.00 & 0.0 $\pm$ 0.00 & 0.0 $\pm$ 0.00 & 0.0 $\pm$ 0.00 \\
\hline
\textbf{Material} & Reward & -6.1 $\pm$ 0.04 & 3.8 $\pm$ 0.49 & 0.4 $\pm$ 0.85 & 0.4 $\pm$ 0.89 & -6.1 $\pm$ 0.04 & 4.2 $\pm$ 0.72 & -0.3 $\pm$ 0.89 & 1.6 $\pm$ 0.67 \\
\textbf{Transport} & Material Left & 68.3 $\pm$ 2.64 & 10.9 $\pm$ 2.28 & 37.2 $\pm$ 9.88 & 26.7 $\pm$ 6.44 & 66.3 $\pm$ 2.64 & 5.5 $\pm$ 2.64 & 35.7 $\pm$ 10.96 & 12.4 $\pm$ 3.22 \\
 & Collisions & 0.0 $\pm$ 0.00 & 0.0 $\pm$ 0.00 & 0.0 $\pm$ 0.00 & 0.0 $\pm$ 0.00 & 0.0 $\pm$ 0.00 & 0.0 $\pm$ 0.00 & 0.0 $\pm$ 0.00 & 0.0 $\pm$ 0.00 \\
\hline
\textbf{Warehouse} & Reward & 6.8 $\pm$ 0.58 & 10.0 $\pm$ 1.48 & 4.0 $\pm$ 0.00 & 5.6 $\pm$ 1.25 & 8.2 $\pm$ 1.30 & 9.2 $\pm$ 1.12 & 3.8 $\pm$ 0.18 & 4.6 $\pm$ 1.03 \\
 & Deliveries & 1.2 $\pm$ 0.18 & 2.2 $\pm$ 0.45 & 0.0 $\pm$ 0.00 & 1.2 $\pm$ 0.31 & 1.6 $\pm$ 0.36 & 1.8 $\pm$ 0.31 & 0.0 $\pm$ 0.00 & 0.8 $\pm$ 0.31 \\
 & Collisions & 0.0 $\pm$ 0.00 & 0.0 $\pm$ 0.00 & 0.0 $\pm$ 0.00 & 0.0 $\pm$ 0.00 & 0.0 $\pm$ 0.00 & 0.0 $\pm$ 0.00 & 0.0 $\pm$ 0.00 & 0.0 $\pm$ 0.00 \\
\hline
\textbf{Navigation} & Reward & -19.5 $\pm$ 2.33 & -14.8 $\pm$ 2.06 & -14.3 $\pm$ 2.64 & -12.8 $\pm$ 2.10 & -17.8 $\pm$ 2.06 & -14.7 $\pm$ 2.01 & -13.7 $\pm$ 2.59 & -12.2 $\pm$ 1.92 \\
\textbf{(MAPF)} & Success Rate & 0.0 $\pm$ 0.00 & 1.0 $\pm$ 0.00 & 1.0 $\pm$ 0.00 & 1.0 $\pm$ 0.00 & 0.2 $\pm$ 0.18 & 1.0 $\pm$ 0.00 & 1.0 $\pm$ 0.00 & 1.0 $\pm$ 0.00 \\
 & Collisions & 0.0 $\pm$ 0.00 & 0.0 $\pm$ 0.00 & 0.0 $\pm$ 0.00 & 0.0 $\pm$ 0.00 & 0.0 $\pm$ 0.00 & 0.0 $\pm$ 0.00 & 0.0 $\pm$ 0.00 & 0.0 $\pm$ 0.00 \\
\hline
\textbf{Foraging} & Reward & 6.2 $\pm$ 1.07 & 6.2 $\pm$ 1.07 & 1.0 $\pm$ 0.89 & 6.6 $\pm$ 2.41 & 5.2 $\pm$ 0.18 & 5.2 $\pm$ 0.18 & 2.0 $\pm$ 1.07 & 11.0 $\pm$ 0.00 \\
 & Resources Foraged & 1.2 $\pm$ 0.18 & 1.2 $\pm$ 0.18 & 0.2 $\pm$ 0.18 & 1.2 $\pm$ 0.45 & 1.0 $\pm$ 0.0 & 1.0 $\pm$ 0.00 & 0.4 $\pm$ 0.22 & 2.0 $\pm$ 0.00 \\
 & Collisions & 0.0 $\pm$ 0.00 & 0.0 $\pm$ 0.00 & 0.0 $\pm$ 0.00 & 0.0 $\pm$ 0.00 & 0.0 $\pm$ 0.00 & 0.0 $\pm$ 0.00 & 0.0 $\pm$ 0.00 & 0.0 $\pm$ 0.00 \\
\hline
\textbf{Predator} & Reward & 44.0 $\pm$ 17.13 & 4.0 $\pm$ 2.19 & 14.0 $\pm$ 8.77 & 52.0 $\pm$ 16.37 & 14.0 $\pm$ 5.37 & 36.0 $\pm$ 16.64 & 6.0 $\pm$ 3.58 & 28.1 $\pm$ 13.64 \\
\textbf{Prey} & Tags & 4.4 $\pm$ 1.70 & 0.4 $\pm$ 0.22 & 1.4 $\pm$ 0.89 & 5.2 $\pm$ 1.65 & 1.4 $\pm$ 0.54 & 3.6 $\pm$ 1.65 & 0.6 $\pm$ 0.36 & 3.8 $\pm$ 1.21 \\
 & Collisions & 0.0 $\pm$ 0.00 & 0.0 $\pm$ 0.00 & 0.0 $\pm$ 0.00 & 0.0 $\pm$ 0.00 & 0.0 $\pm$ 0.00 & 0.0 $\pm$ 0.00 & 0.0 $\pm$ 0.00 & 0.0 $\pm$ 0.00 \\
\hline
\textbf{Continuous} & Reward & 1.0 $\pm$ 0.67 & 0.8 $\pm$ 0.54 & 0.0 $\pm$ 0.00 & 0.0 $\pm$ 0.00 & 0.8 $\pm$ 0.31 & 6.2 $\pm$ 2.68 & 0.0 $\pm$ 0.00 & 0.0 $\pm$ 0.00 \\
\textbf{RWARE} & Shelf Dropoffs & 1.0 $\pm$ 0.67 & 0.8 $\pm$ 0.54 & 0.0 $\pm$ 0.00 & 0.0 $\pm$ 0.00 & 0.8 $\pm$ 0.31 & 6.2 $\pm$ 2.68 & 0.0 $\pm$ 0.00 & 0.0 $\pm$ 0.00 \\
 & Collisions & 0.0 $\pm$ 0.00 & 0.0 $\pm$ 0.00 & 0.0 $\pm$ 0.00 & 0.0 $\pm$ 0.00 & 0.0 $\pm$ 0.00 & 0.0 $\pm$ 0.00 & 0.0 $\pm$ 0.00 & 0.0 $\pm$ 0.00 \\
\hline
\end{tabular}%
}
\caption{Performance metrics across scenarios for simulated and real-world experiments.}
\label{tab:sim2real}
\end{table*}

\textbf{Sim2Real}: As one would expect, we found reasonable gaps between the simulated and real world runs for most tasks across methods (see Table \ref{tab:sim2real}). Some of this gap could be attributed to only having limited real-world runs and the stochasticity of the real world compared to the deterministic simulator. However, two of our new tasks (Predator Prey and Continuous-RWARE) resulted in surprisingly large sim2real performance gaps, particularly for QMIX and PQN. Our qualitative analysis of the rollouts revealed that these tasks contain inflection points that can exacerbate modest differences into large differences in the environment and robot behavior over time. 
For instance, small positional changes to the robots in Predator-Prey can sometimes elicit qualitatively different prey behaviors which cascade into dramatic differences between the simulated and real-world rollouts.
% Specifically, we observed that the simulated and real-world rollouts would initially align, but small differences in robot positions would sometimes dramatically alter the interaction between robots and the prey/shelf. 
% Following this inflection point, the simulation and reality diverged, leading to drastically different outcomes. 
In contrast, existing MARBLER tasks are less sensitive to such variations, as robots mostly interact with a static environment. The observed gaps in the new tasks underscore the importance of hardware evaluation, especially in multi-robot scenarios where task outcomes are closely tied to interaction dynamics. The ability to accurately simulate single robots does not necessarily imply accurate multi-robot simulations.

\begin{wraptable}{r}{0.7\textwidth}
\vspace{-1em}
\resizebox{0.7\textwidth}{!}{%
\begin{tabular}{ll|cc|cc}
\textbf{Scenario} & \textbf{Metric} & \multicolumn{2}{c|}{\textbf{Simulated Experiments}} & \multicolumn{2}{c}{\textbf{Real-World Experiments}} \\
& & \textbf{PQN} & \textbf{QMIX} & \textbf{PQN} & \textbf{QMIX} \\
\hline
\textbf{Predator Prey} & Returns & 30.0 $\pm$ 6.33 & 3.3 $\pm$ 1.21 & 16.0 $\pm$ 5.06 & 26.0 $\pm$ 11.33 \\
& Tags & 3.0 $\pm$ 0.62 & 0.3 $\pm$ 0.13 & 1.6 $\pm$ 0.52 & 2.6 $\pm$ 1.14 \\
& Collisions & 0.0 $\pm$ 0.00 & 0.0 $\pm$ 0.00 & 0.0 $\pm$ 0.00 & 0.0 $\pm$ 0.00 \\
\hline
\textbf{Predator Prey} & Returns & 66.7 $\pm$ 3.23 & 16.7 $\pm$ 6.09 & 46.7 $\pm$ 9.66 & 18.0 $\pm$ 4.60 \\
\textbf{w/ DR} & Tags & 6.7 $\pm$ 0.34 & 1.7 $\pm$ 0.62 & 4.7 $\pm$ 0.96 & 1.8 $\pm$ 0.46 \\
& Collisions & 0.0 $\pm$ 0.00 & 0.0 $\pm$ 0.00 & 0.0 $\pm$ 0.00 & 0.0 $\pm$ 0.00 \\
\hline
\end{tabular}%
}
\caption{Predator Prey performance with and without domain randomization (3 initial conditions, 5 trials per initial condition).}
\label{tab:pred-prey-dr}
\vspace{-1em}
\end{wraptable}

\textbf{Domain Randomization}: Inspired by the above observations, we applied simple domain randomization through action noise to learn a more robust policy. We found that training with action noise can reduce the sim2real gap in Predator Prey for both QMIX and PQN and improve task performance in simulation (see \tableautorefname~\ref{tab:pred-prey-dr}).

\begin{figure}
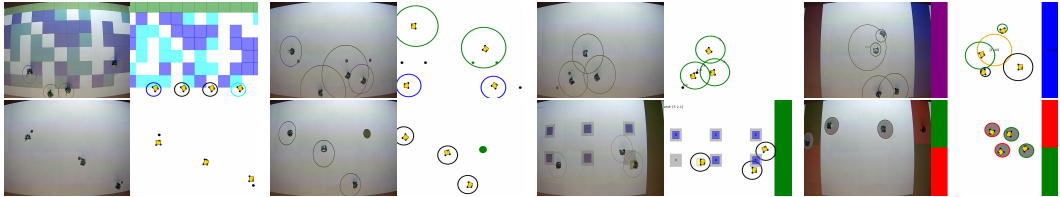

    \centering
    \foreach \img in {at, discovery, foraging, mt, navigation, pp, rware, warehouse} {
        \begin{subfigure}{0.24\columnwidth}
            \includegraphics[width=0.5\columnwidth,height=0.375\columnwidth,keepaspectratio=false,trim=50 20 50 20,clip]{figures/deployment-pqn/\img.png}%
            \includegraphics[width=0.5\columnwidth,height=0.375\columnwidth,keepaspectratio=false,trim=50 20 50 20,clip]{figures/deployment-pqn/\img_sim.png}
        \end{subfigure}
    }
    \caption{Scenarios (real on left and sim on right) from left to right and top to bottom as follows: Arctic Transport, Discovery, Foraging, Material Transport, Navigation, Predator Prey, RWARE, Warehouse. For detailed descriptions of each scenario, please see Appendix \ref{appendix:scenario-details}.}
    \label{fig:sim2real}
\end{figure}
\vspace{-1em}

%===============================================================================
\section{Conclusion}
We contribute JaxRobotarium, an accelerated benchmark platform for MRRL with free, publicly available, and standardized sim2real evaluation through the Robotarium hardware test bed. Our benchmark experiments demonstrate up to a 150x speed up in simulation, and a 20x speed up in training multi-robot policies compared to baseline. We demonstrate the utility of our platform by evaluating 4 MARL algorithms on 8 multi-robot coordination scenarios, with a large-scale sim2real evaluation consisting of over 200 real world deployments. We hope JaxRobotarium will be a valuable resource for accelerating the development, benchmarking, and sim2real evaluation of MRRL algorithms and architectures.

% \section{Instructions}
% Submission to CoRL 2025 will be entirely electronic, via a web site (not email). Information about the submission process and \LaTeX{} templates are available on the conference web site at \url{https://corl.org/}. For camera ready submission, use the \texttt{final} option for the \verb|\usepackage| command. 

% \section{Citations}
% \label{sec:citations}

% 	Citations can be made using either \textbackslash citep\{\} or \textbackslash citet\{\}, depending from the appropriateness. To avoid the citation moving to the next line, it is often a good practice to replace the space before with a tilde (\~{}) character.
% 	Example 1: ``CoRL is the best conference ever~\citep{Gauss1857}.''
% 	Example 2: ``\citet{Lagrange1788} proved, both theoretically and numerically, that CoRL is the best conference ever.''
	
%===============================================================================
\clearpage
\section{Limitations}
We design our platform for sim2real evaluation through the Robotarium testbed, so we only simulate the Robotarium GRITSBots and we do not simulate observations from more complex perception modules, such as images or LiDAR scans.

The decision to utilize the Robotarium introduces additional limitations. Since the robots are homogeneous, heterogeneity has to be artificially injected, such as limiting a robot's velocity or waypoint step size, or through introducing scenario-specific heterogeneity that is recognized by the scenario's step function, such as obeying the foraging level of robots when deciding if a resource is successfully foraged. Furthermore, the Robotarium hardware test bed does not support Jax libraries, so the quadratic problem solver used for barrier certificates in deployment and the solver used in Jax simulation (chosen differently for compatibility with \texttt{jit} and \texttt{vmap}) are not aligned. We observe that underlying differences in the solvers can result in different proposed solutions to the same quadratic program.

We note that our platform cannot be considered a replacement for training in high-fidelity robotic simulation environments such as IsaacGym \cite{makoviychuk2021isaac} and IsaacLab \cite{mittal2023orbit}. Rather, JaxRobotarium is designed to be a turn-key platform for simulation-training-deployment tailored for MRRL research, offering native integrations (e.g., JaxMARL), standardized benchmarks (e.g., RWARE adaptations), and a ready-to-use zero-cost sim2real pipeline.

Finally, in our training evaluations, we only use an RNN-based policy architecture since it is widely implemented to overcome the challenges of partial observability and long-horizon reasoning \cite{qmix, mappo-ippo, jaxmarl, papoudakis2021benchmarking, torbati2023marbler}, and do not evaluate on MLP architectures \cite{bettini2022vmas, bettini2024benchmarl}, GNN architectures \cite{li2021magat, howell2024generalization, bettini2023heterogeneous, kailas2025evaluating}, or more recent Hyper-Network architectures \cite{fu2025learning, tessera2025hypermarl}. Additionally, we train with limited partial observability in most scenarios, see Appendix \ref{appendix:scenario-details} for exact details. Future work could benchmark these policy architectures within \frameworkName{} with stronger partial observability.

%===============================================================================
% \clearpage
% % The acknowledgments are automatically included only in the final and preprint versions of the paper.
% \acknowledgments{If a paper is accepted, the final camera-ready version will (and probably should) include acknowledgments. All acknowledgments go at the end of the paper, including thanks to reviewers who gave useful comments, to colleagues who contributed to the ideas, and to funding agencies and corporate sponsors that provided financial support.}

%===============================================================================

\bibliography{main}  % .bib

\clearpage
\appendix
\begin{figure}[t]
    \centering
    \includegraphics[width=\columnwidth]{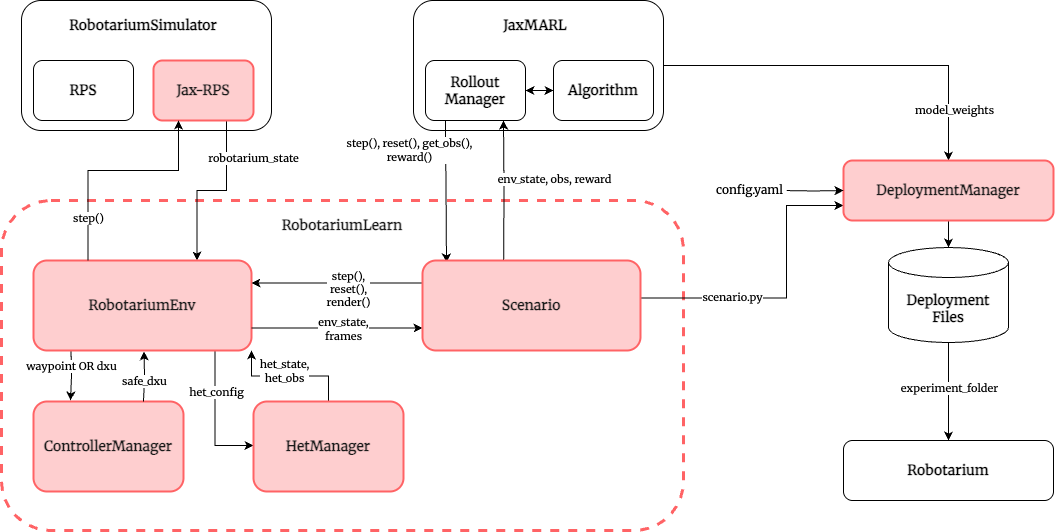}
    \caption{\frameworkName{} architecture, colored components are novel.}
    \label{fig:block_diagram}
\end{figure}

%===============================================================================
\section{Training Details}
\label{appendix:training-details}
\subsection{Policy Architecture}
We use an RNN-based policy architecture across all of our training experiments. The specific implementations are all adopted from JaxMARL's baseline implementations \cite{jaxmarl} and use a Gated Recurrent Unit (GRU) \cite{cho2014learningphraserepresentationsusing}.
\begin{flushitemize}
    \item PQN: A 2 layer MLP with Layer Normalization before ReLU activations, followed by a GRU, followed by a linear layer.
    \item QMIX: A 1 layer MLP with ReLU activation, followed by a GRU, followed by a linear layer.
    \item MAPPO: A single layer MLP with ReLU activation, followed by a GRU, followed by a single layer MLP with ReLU activation, followed by a linear layer.
    \item IPPO: A single layer MLP with ReLU activation, followed by a GRU, followed by a single layer MLP with ReLU activation, followed by a linear layer.
\end{flushitemize}
For all algorithms, we train 5 policies in parallel across 5 unique seeds. Exact widths for the policies' hidden layers were chosen by ablating over the widths 128, 256, and 512 trained on a single seed. The best-performing width was then chosen for the final training runs. We report all policy widths in Table \ref{tab:policy-hidden-width}.

\begin{table*}[ht]
\centering
\resizebox{0.5\textwidth}{!}{%
\begin{tabular}{l|cccc}

\textbf{Scenario} & \textbf{PQN} & \textbf{QMIX} & \textbf{MAPPO} & \textbf{IPPO} \\
\hline

\textbf{Arctic} & 256 & 256 & 128 & 128 \\
\textbf{Transport} & & & & \\
\hline
\textbf{Discovery} & 256 & 256 & 128 & 256 \\
\hline
\textbf{Material} & 512 & 128 & 128 & 512 \\
\textbf{Transport} & & & & \\
\hline
\textbf{Warehouse} & 256 & 512 & 128 & 256 \\
\hline
\textbf{Navigation} & 256 & 512 & 128 & 512 \\
\hline
\textbf{Foraging} & 128 & 128 & 512 & 256 \\
\hline
\textbf{Predator} & 128 & 512 & 256 & 512 \\
\textbf{Prey} & & & & \\
\hline
\textbf{Continuous} & 256 & 256 & 128 & 256 \\
\textbf{RWARE} & & & & \\
\hline

\end{tabular}%
}
\caption{Policy hidden width across scenarios and algorithms.}
\label{tab:policy-hidden-width}
\end{table*}

\subsection{PQN}
\begin{table*}[h]
\centering
\begin{tabular}{l|c}
\textbf{Parameter} & \textbf{Value} \\
\hline
NUM\_ENVS & 256 \\
NORM\_INPUTS & False \\
NORM\_TYPE & layer\_norm \\
EPS\_START & 1.0 \\
EPS\_FINISH & 0.1 \\
EPS\_DECAY & 0.1 \\
MAX\_GRAD\_NORM & 1 \\
NUM\_MINIBATCHES & 16 \\
NUM\_EPOCHS & 4 \\
LR & 0.0005 \\
LR\_LINEAR\_DECAY & False \\
GAMMA & 0.9 \\
LAMBDA & 0.3 \\
\end{tabular}
\caption{PQN hyperparameters}
\label{tab:pqn-hyperparams}
\end{table*}
We train with the JaxMARL tuned hyperparameters listed in Table \ref{tab:pqn-hyperparams}.

\subsection{QMIX}
\begin{table*}[h]
\centering
\begin{tabular}{l|c}
\textbf{Parameter} & \textbf{Value} \\
\hline
NUM\_ENVS & 8 \\
BUFFER\_SIZE & 5000 \\
BUFFER\_BATCH\_SIZE & 32 \\
MIXER\_EMBEDDING\_DIM & 32 \\
MIXER\_HYPERNET\_HIDDEN\_DIM & 128 \\
MIXER\_INIT\_SCALE & 0.001 \\
EPS\_START & 1.0 \\
EPS\_FINISH & 0.05 \\
EPS\_DECAY & 0.1 \\
MAX\_GRAD\_NORM & 25 \\
TARGET\_UPDATE\_INTERVAL & 200 \\
TAU & 1.0 \\
NUM\_EPOCHS & 1 \\
LR & 0.005 \\
LEARNING\_STARTS & 10000 \\
LR\_LINEAR\_DECAY & True \\
GAMMA & 0.9 \\
\end{tabular}
\caption{QMIX hyperparameters}
\label{tab:qmix-hyperparams}
\end{table*}

We train with the JaxMARL tuned hyperparameters listed in Table \ref{tab:qmix-hyperparams}.

\subsection{MAPPO}
\begin{table*}[h]
\centering
\begin{tabular}{l|c}
\textbf{Parameter} & \textbf{Value} \\
\hline
LR & 0.002 \\
NUM\_ENVS & 16 \\
HIDDEN\_SIZE & 128 \\
UPDATE\_EPOCHS & 4 \\
NUM\_MINIBATCHES & 4 \\
GAMMA & 0.99 \\
GAE\_LAMBDA & 0.95 \\
CLIP\_EPS & 0.2 \\
SCALE\_CLIP\_EPS & False \\
ENT\_COEF & 0.01 \\
VF\_COEF & 0.5 \\
MAX\_GRAD\_NORM & 0.5 \\
ANNEAL\_LR & True \\
\end{tabular}
\caption{MAPPO hyperparameters}
\label{tab:mappo-hyperparams}
\end{table*}

We train with the JaxMARL tuned hyperparameters listed in Table \ref{tab:mappo-hyperparams}.

\subsection{IPPO}
\begin{table*}[h]
\centering
\begin{tabular}{l|c}
\textbf{Parameter} & \textbf{Value} \\
\hline
LR & 0.0005 \\
NUM\_ENVS & 16 \\
UPDATE\_EPOCHS & 4 \\
NUM\_MINIBATCHES & 2 \\
GAMMA & 0.99 \\
GAE\_LAMBDA & 0.95 \\
CLIP\_EPS & 0.3 \\
SCALE\_CLIP\_EPS & False \\
ENT\_COEF & 0.01 \\
VF\_COEF & 1.0 \\
MAX\_GRAD\_NORM & 0.5 \\
ANNEAL\_LR & True \\
\end{tabular}
\caption{IPPO hyperparameters}
\label{tab:ippo-hyperparams}
\end{table*}

We train with the JaxMARL tuned hyperparameters listed in Table \ref{tab:ippo-hyperparams}.

\begin{figure}[t!b]
    \centering
    \begin{subfigure}{0.24\columnwidth}
      \includegraphics[width=\columnwidth]{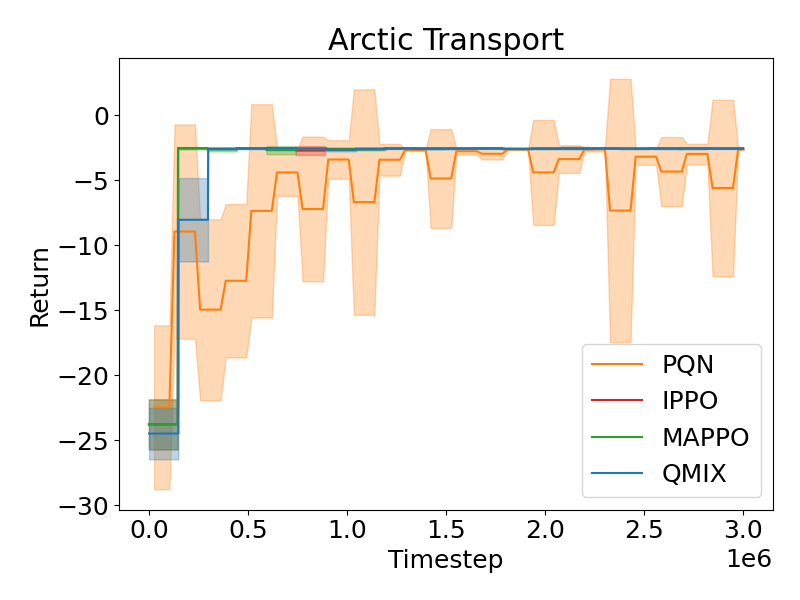}
    \end{subfigure}
    \begin{subfigure}{0.24\columnwidth}
      \includegraphics[width=\columnwidth]{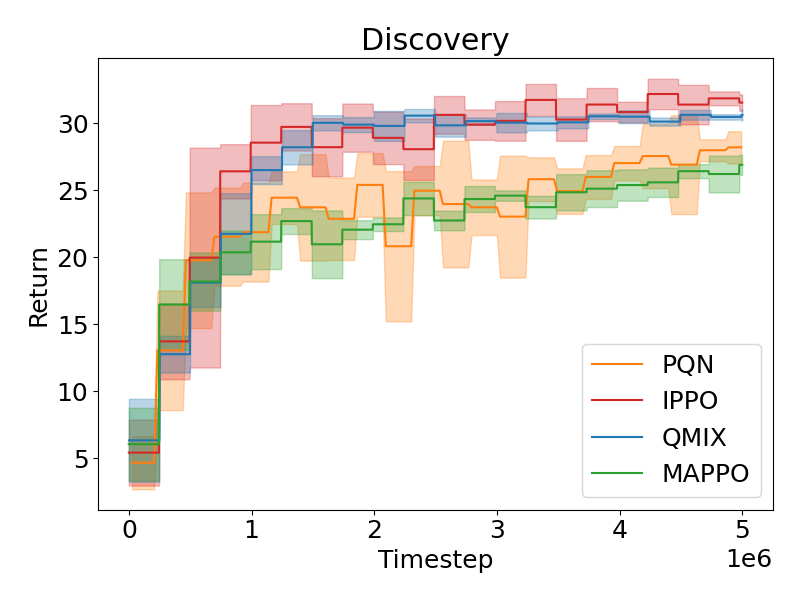}
    \end{subfigure}
    \begin{subfigure}{0.24\columnwidth}
      \includegraphics[width=\columnwidth]{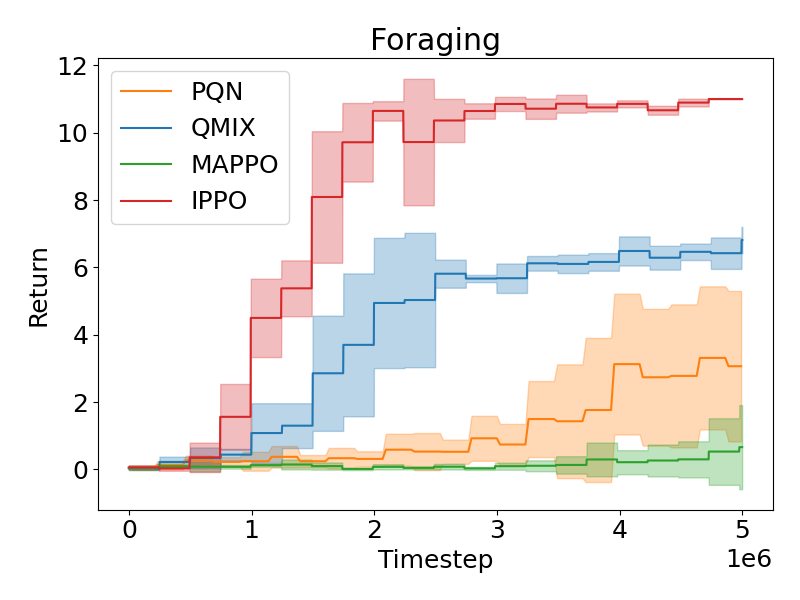}
    \end{subfigure}
    \begin{subfigure}{0.24\columnwidth}
      \includegraphics[width=\columnwidth]{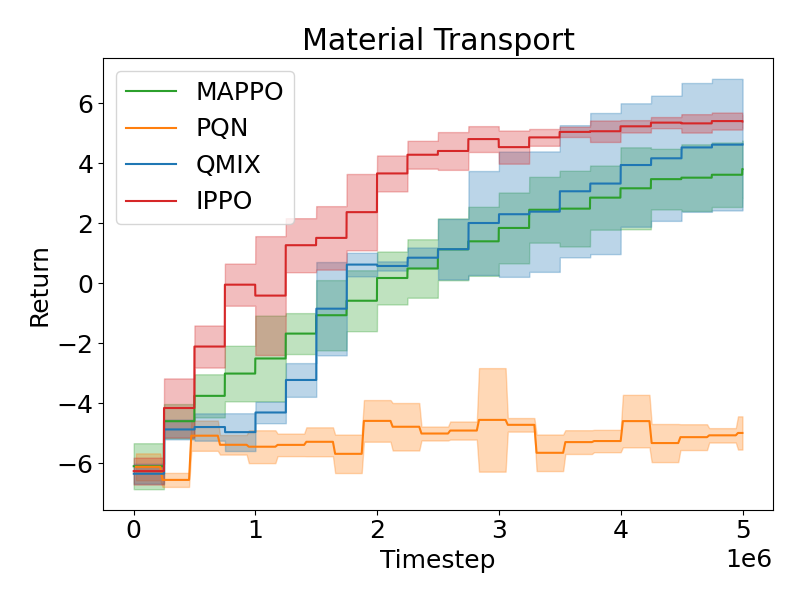}
    \end{subfigure}

    \begin{subfigure}{0.24\columnwidth}
      \includegraphics[width=\columnwidth]{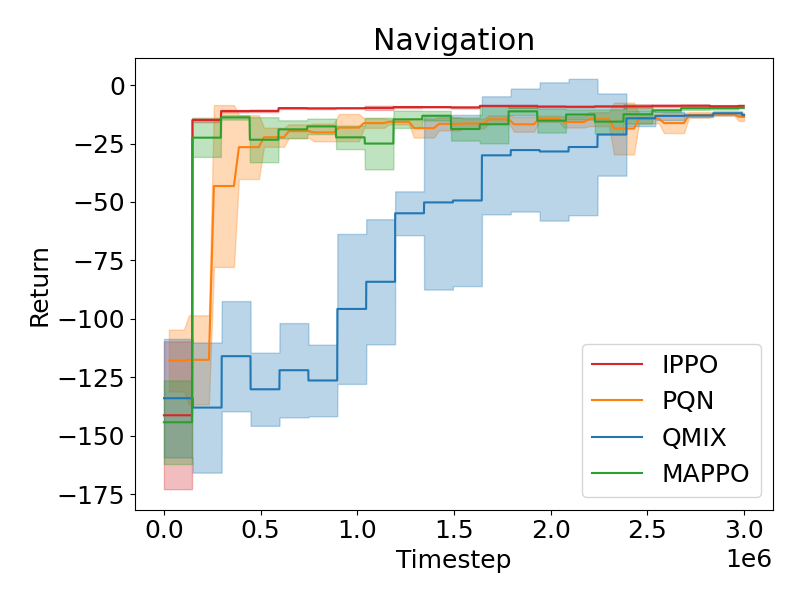}
    \end{subfigure}
    \begin{subfigure}{0.24\columnwidth}
      \includegraphics[width=\columnwidth]{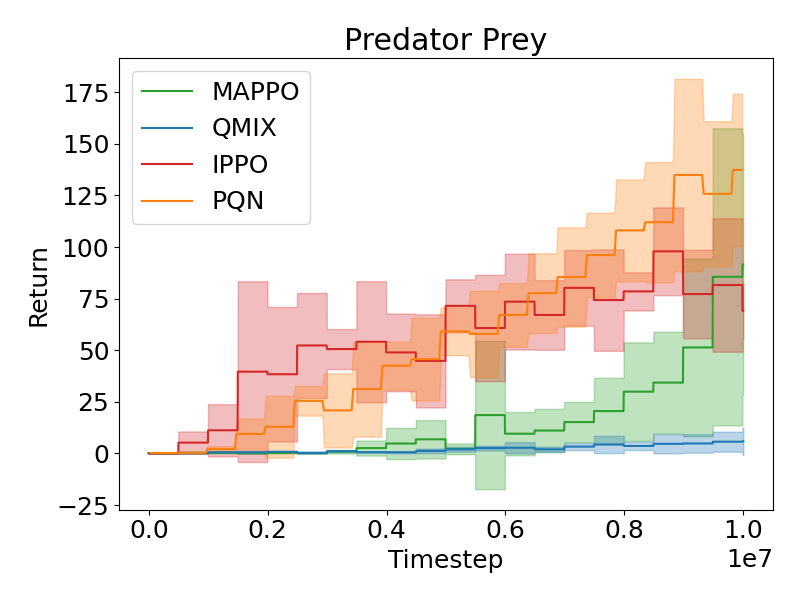}
    \end{subfigure}
    \begin{subfigure}{0.24\columnwidth}
      \includegraphics[width=\columnwidth]{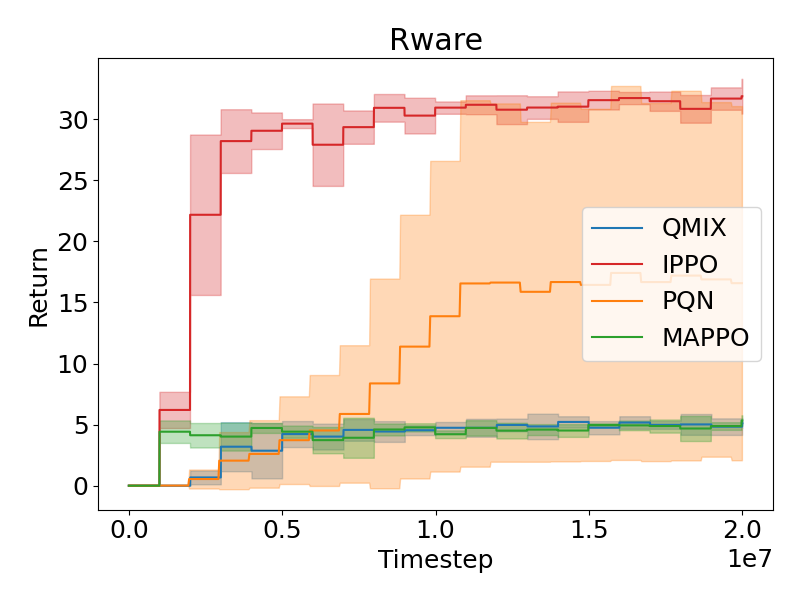}
    \end{subfigure}
    \begin{subfigure}{0.24\columnwidth}
      \includegraphics[width=\columnwidth]{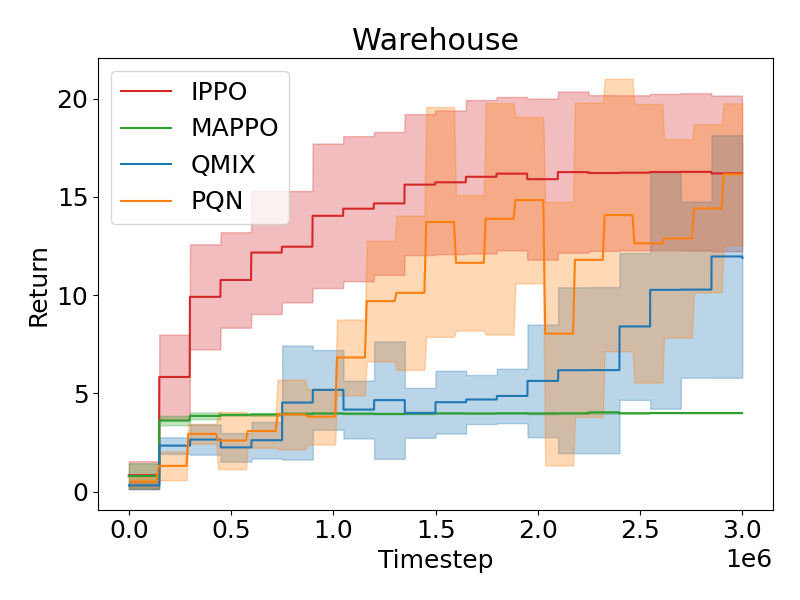}
    \end{subfigure}
    
    \caption{We plot the greedy returns throughout training, where all algorithms are trained for the same number of timesteps for each scenario}
    \label{fig:train-returns}
\end{figure}

%===============================================================================
\section{Scenario Details}
\label{appendix:scenario-details}
As a general note for all scenarios, if heterogeneity awareness is configured, robots will additionally observe the configured heterogeneity representation. In our training experiments, the action space for all robots is waypoints that are followed by a unicycle position controller with barrier functions enabled for collision avoidance.

\subsection{Arctic Transport}
\begin{figure}[h]
    \centering
    \begin{subfigure}{0.48\columnwidth}
        \includegraphics[width=\columnwidth,height=0.75\columnwidth,keepaspectratio=false]{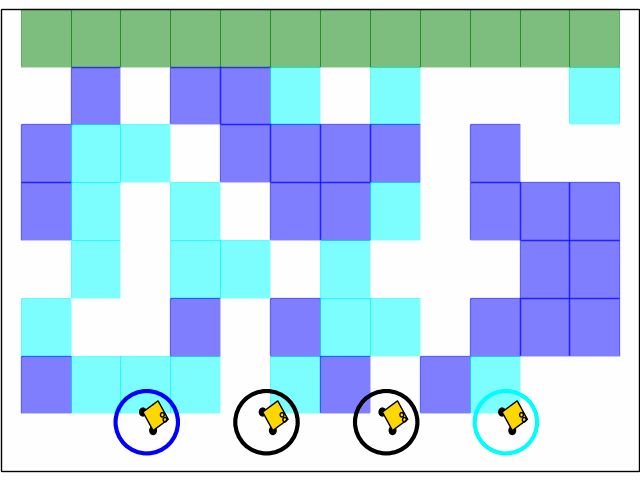}
    \end{subfigure}
    \begin{subfigure}{0.48\columnwidth}
        \includegraphics[width=\columnwidth,height=0.75\columnwidth,keepaspectratio=false]{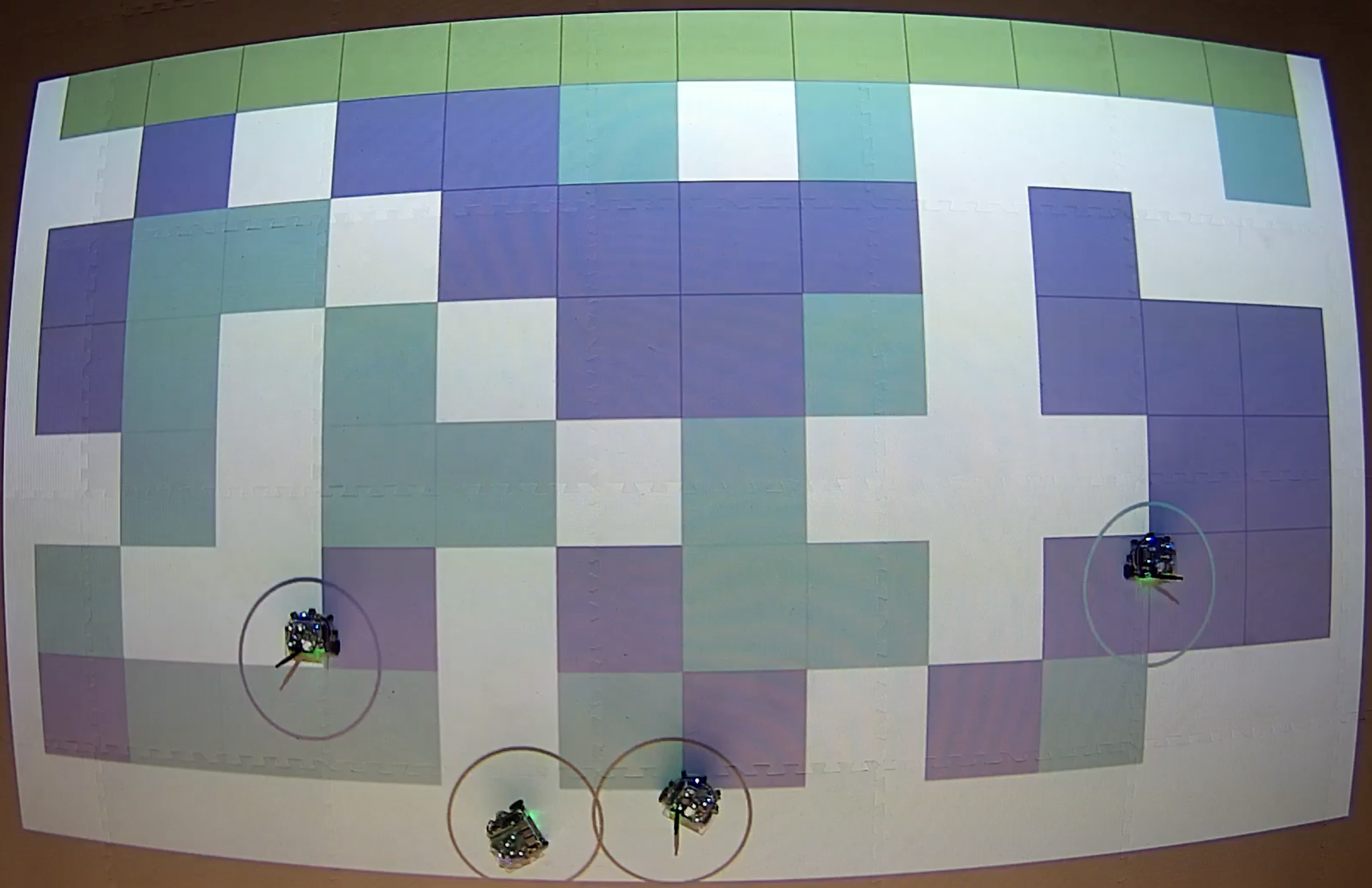}
    \end{subfigure}
    \caption{Arctic Transport in simulation (left) and real-world (right).}
    \label{fig:at-sim-real}
\end{figure}

\subsubsection{Description}
Adopted from MARBLER, in this scenario, the ice and water robots must cross the terrain consisting of ice tiles (light blue), water tiles (dark blue), and ground tiles (white) to reach the goal zone (green). Ice and water robots move at the same speed across ground tiles, slow over tiles not corresponding to their type, and fast over tiles corresponding to their types. Ice and water robots observe their position, the position of other robots, and the current tile type they are occupying. Drone robots observe a 3x3 grid centered around their current position, and move at a uniform speed across all tiles. Their observations are directly communicated to the ice and water robots, such that they can help guide the ice and water robots. The team is rewarded for minimizing the distance of the ice and water robots to the goal zone, and penalized for every step taken where both robots are not in the goal zone.

\subsubsection{Experiment Parameters}
\begin{flushitemize}
    \item Number of Robots: 4
    \item Max steps: 100
    \item Training timesteps: 3e6
    \item Controller
    \begin{flushitemize}
        \item Controller: unicycle Position
        \item Barrier Fn: enabled
    \end{flushitemize}
    \item Normal Step: 0.2
    \item Fast Step: 0.3
    \item Slow Step: 0.1
    \item Heterogeneity
    \begin{flushitemize}
        \item Type: class
        \item Obs Type: class (robots observe their own class)
        \item Values: [[1, 0, 0], [1, 0, 0], [0, 1, 0], [0, 0, 1]] (each entry indicates the type [drone, water, ice])
    \end{flushitemize}
    \item Distance Reward: -0.05 $\times$ summed shortest distances of ice and water robots to the goal zone
    \item Step Reward: -0.10 $\times$ 0 if all robots are on their goal, otherwise 1
    \item Violation Reward: 0 $\times$ number of collisions
\end{flushitemize}

\subsection{Discovery}
\begin{figure}[h]
    \centering
    \begin{subfigure}{0.48\columnwidth}
        \includegraphics[width=\columnwidth,height=0.75\columnwidth,keepaspectratio=false]{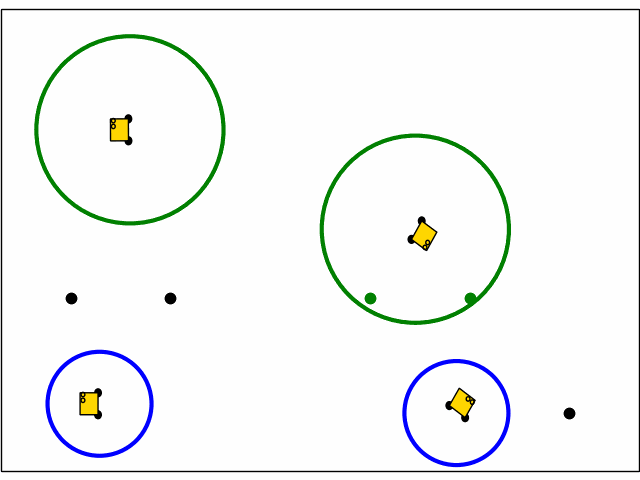}
    \end{subfigure}
    \begin{subfigure}{0.48\columnwidth}
        \includegraphics[width=\columnwidth,height=0.75\columnwidth,keepaspectratio=false]{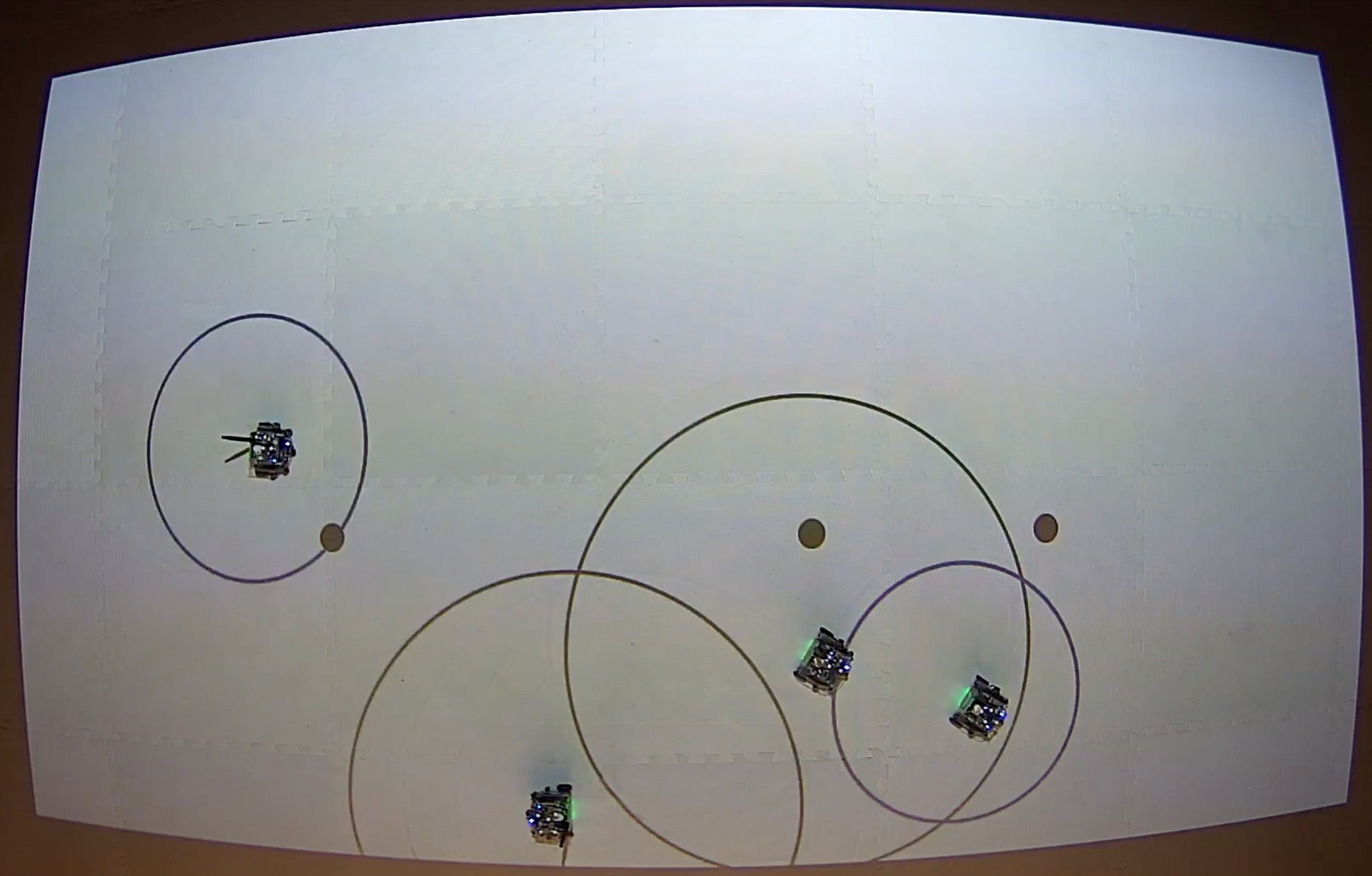}
    \end{subfigure}
    \caption{Discovery in simulation (left) and real-world (right).}
    \label{fig:discovery-sim-real}
\end{figure}

\subsubsection{Description}
Renamed from Predator-Capture-Prey in MARBLER to better reflect the scenario objective, in this scenario, N sensing robots must collaborate with M tagging robots to tag K landmarks (black) in the environment. Sensing robots and tagging robots have configurable sensing radii. Robots observe their own location and the location of other robots. Initially, robot observations also contain a position outside the map for all landmarks, indicating they have not been sensed. When a landmark is within range of a sensing robot, it is considered sensed, the black marker is updated to green, and the location becomes available within the observation. When a landmark is within range of a tagging robot, the landmark is considered tagged and disappears from the visualization. The team is rewarded for landmarks sensed and tagged, and penalized for every step taken where all landmarks are not tagged. This scenario can scale to an arbitrary number of robots and landmarks.

\subsubsection{Experiment Parameters}
\begin{flushitemize}
    \item Number of Robots: 4
    \item Sensing Robots: 2
    \item Tagging Robots: 2
    \item Max Steps: 100
    \item Training timesteps: 5e6
    \item Controller
    \begin{flushitemize}
        \item Controller: unicycle position
        \item Barrier Fn: enabled
    \end{flushitemize}
    \item Heterogeneity
    \begin{flushitemize}
        \item Type: capability set
        \item Obs type: full capability set (robots observe the team's capability values)
        \item Values: [[0.45, 0], [0.45, 0], [0, 0.25], [0, 0.25]] (each entry is [sensing radius, tag radius])
    \end{flushitemize}
    \item Sensing Reward: 1 $\times$ landmarks sensed
    \item Tagging Reward: 5 $\times$ landmarks tagged
    \item Step Reward: -0.05 $\times$ 0 if all landmarks tagged, otherwise 1
    \item Violation Reward: 0 $\times$ number of collisions
\end{flushitemize}

\subsection{Foraging}
\begin{figure}[h]
    \centering
    \begin{subfigure}{0.48\columnwidth}
        \includegraphics[width=\columnwidth,height=0.75\columnwidth,keepaspectratio=false]{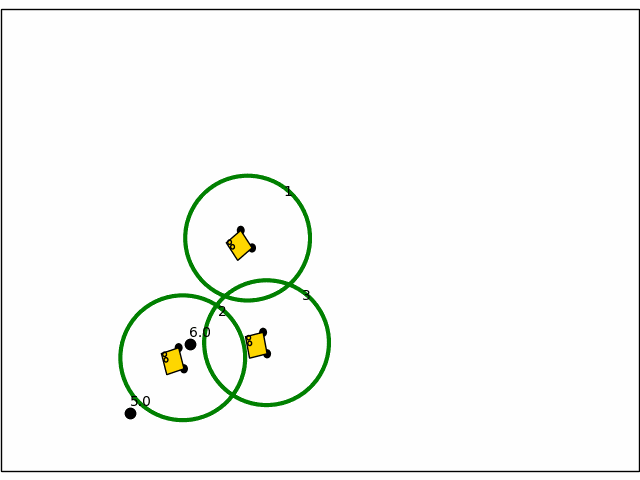}
    \end{subfigure}
    \begin{subfigure}{0.48\columnwidth}
        \includegraphics[width=\columnwidth,height=0.75\columnwidth,keepaspectratio=false]{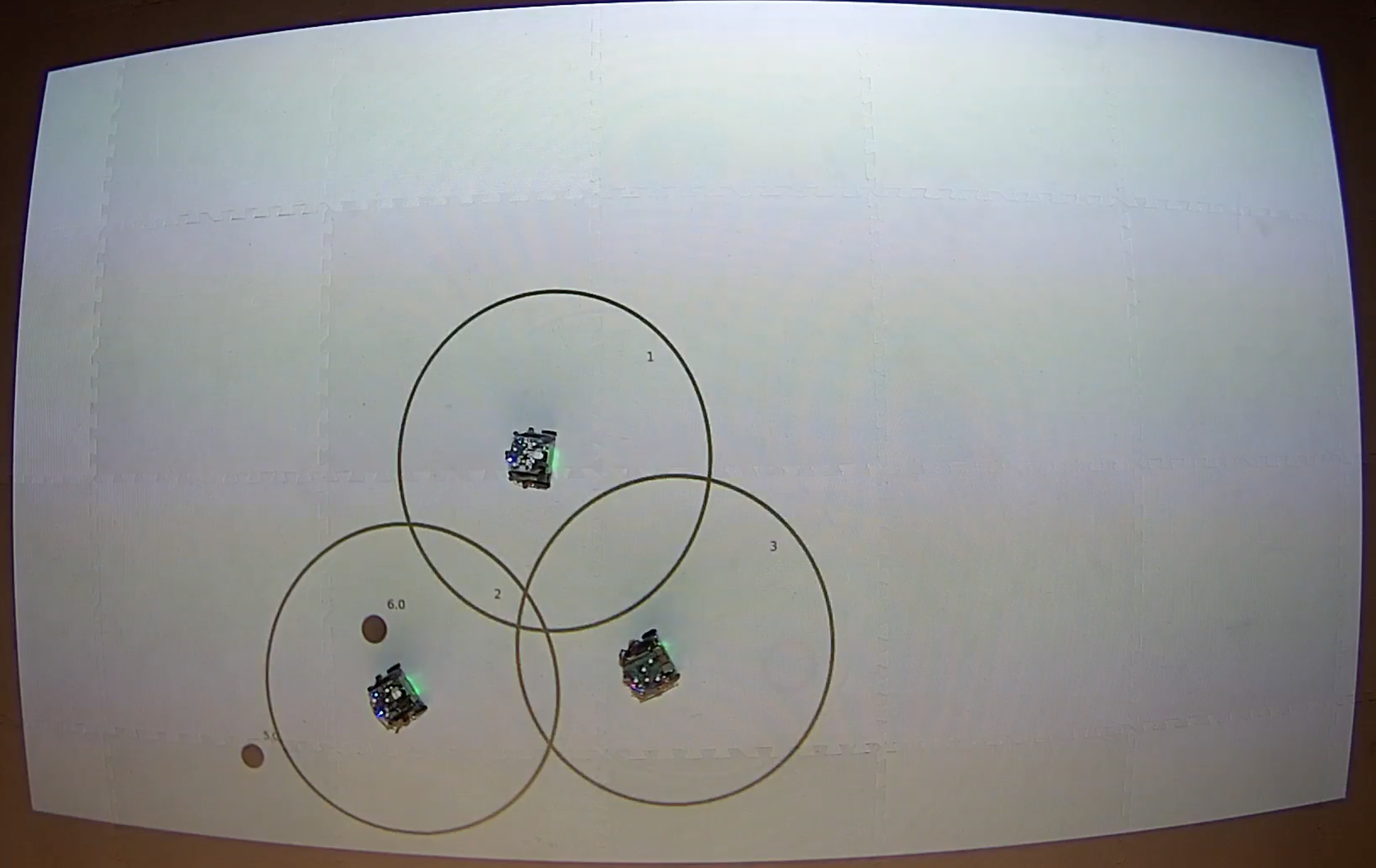}
    \end{subfigure}
    \caption{Foraging in simulation (left) and real-world (right).}
    \label{fig:foraging-sim-real}
\end{figure}

\subsubsection{Description}
Inspired by grid-world level-based foraging in \cite{lbf-ref}, in this scenario, N robots with varying levels must collaborate to forage M resources (black markers) within the environment. Labels indicate the levels of robots and resources. Robots observe their position, the position of other robots, the location of resources, and their corresponding levels. For a resource to be considered foraged, it must be within the foraging radius of M robots, where the summed level of the M robots is greater than or equal to the level of the resource. When resources are foraged, they disappear from the visualization to indicate they have been collected. The team is rewarded for each resource foraged, scaled by the level of that resource. This scenario can scale to an arbitrary number of robots and resources.

\subsubsection{Experiment Parameters}
\begin{flushitemize}
    \item Number of Robots: 3
    \item Number of Resources: 2
    \item Max steps: 100
    \item Training timesteps: 5e6
    \item Controller
    \begin{flushitemize}
        \item Controller: unicycle position
        \item Barrier Fn: enabled
    \end{flushitemize}
    \item Heterogeneity
    \begin{flushitemize}
        \item Type: capability set
        \item Obs type: full capability set (robots observe the team's capability values)
        \item Values: [[1], [2], [3]] (each entry is the robot's foraging level)
    \end{flushitemize}
    \item Foraging Reward: 1 $\times$ level of resources foraged
    \item Violation Reward: 0 $\times$ number of collisions
\end{flushitemize}

\subsection{Material Transport}
\begin{figure}[h]
    \centering
    \begin{subfigure}{0.48\columnwidth}
        \includegraphics[width=\columnwidth,height=0.75\columnwidth,keepaspectratio=false]{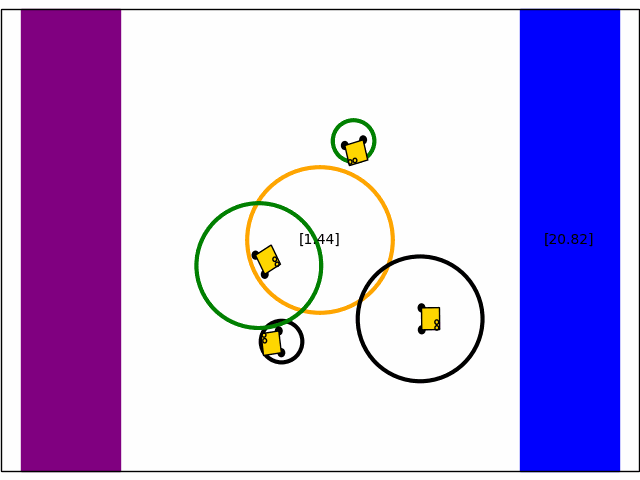}
    \end{subfigure}
    \begin{subfigure}{0.48\columnwidth}
        \includegraphics[width=\columnwidth,height=0.75\columnwidth,keepaspectratio=false]{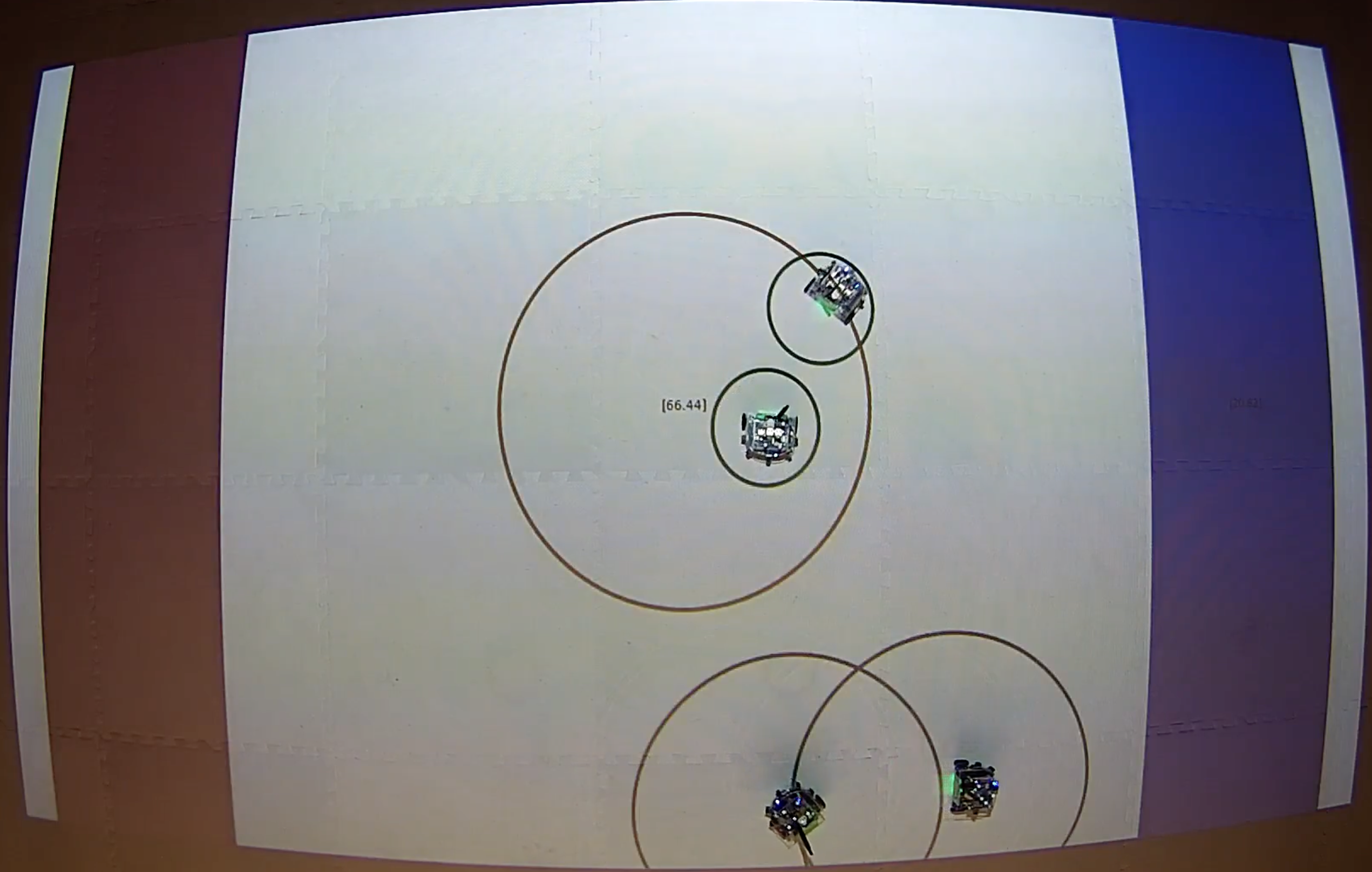}
    \end{subfigure}
    \caption{Material Transport in simulation (left) and real-world (right).}
    \label{fig:mt-sim-real}
\end{figure}

\subsubsection{Description}
Adopted from MARBLER, in this scenario, N robots with varying speeds and carrying capacities must collaborate to unload two loading zones (green circle and green rectangle) into the drop-off zone (purple rectangle). Both zones are initialized with an amount of material sampled from a zone-specific distribution. We assume both distributions are Gaussian with pre-specified means and variances. Robots observe their position, the position of other robots, and the remaining material to unload in both zones. Robots can only load material by entering the loading zones, can only carry the amount of material specified to their carrying capacity, and must drop off their loaded material into the drop-off zone before loading more material. The team is rewarded for the amount of material loaded and the amount of material dropped off, and it is penalized for each step taken where there is material remaining in the loading zones. This scenario can scale to an arbitrary number of robots.

\subsubsection{Experiment Parameters}
\begin{flushitemize}
    \item Number of Robots: 4
    \item Max steps: 70
    \item Training timesteps: 5e6
    \item Controller
    \begin{flushitemize}
        \item Controller: unicycle position
        \item Barrier Fn: enabled
    \end{flushitemize}
    \item Heterogeneity
    \begin{flushitemize}
        \item Type: capability set
        \item Obs type: full capability set (robots observe the team's capability values)
        \item Values: [[0.45, 5], [0.45, 5], [0.15, 15], [0.15, 15]] (each entry is [step size, carrying capacity])
    \end{flushitemize}
    \item Gaussian distribution parameters of materials in circular loading zone
    \begin{flushitemize}
        \item Mean: 75
        \item Variance: 10
    \end{flushitemize}
    \item Gaussian distribution parameters of materials in rectangular loading zone
    \begin{flushitemize}
        \item Mean: 15
        \item Variance: 4
    \end{flushitemize}
    \item Load Reward: 0.25 $\times$ material loaded by robots
    \item Drop Off Reward: 0.75 $\times$ material dropped off by robots
    \item Step Reward: -0.1 $\times$ 0 if no material remaining in loading zones, otherwise 1
    \item Violation Reward: 0 $\times$ number of collisions
\end{flushitemize}

\subsection{Navigation (MAPF)}
\begin{figure}[h]
    \centering
    \begin{subfigure}{0.48\columnwidth}
        \includegraphics[width=\columnwidth,height=0.75\columnwidth,keepaspectratio=false]{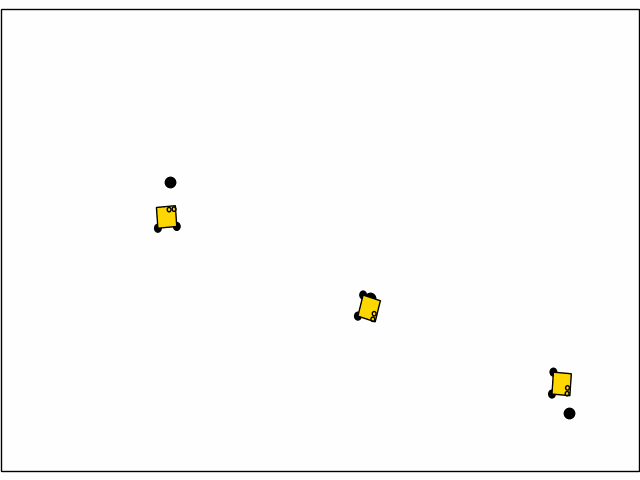}
    \end{subfigure}
    \begin{subfigure}{0.48\columnwidth}
        \includegraphics[width=\columnwidth,height=0.75\columnwidth,keepaspectratio=false]{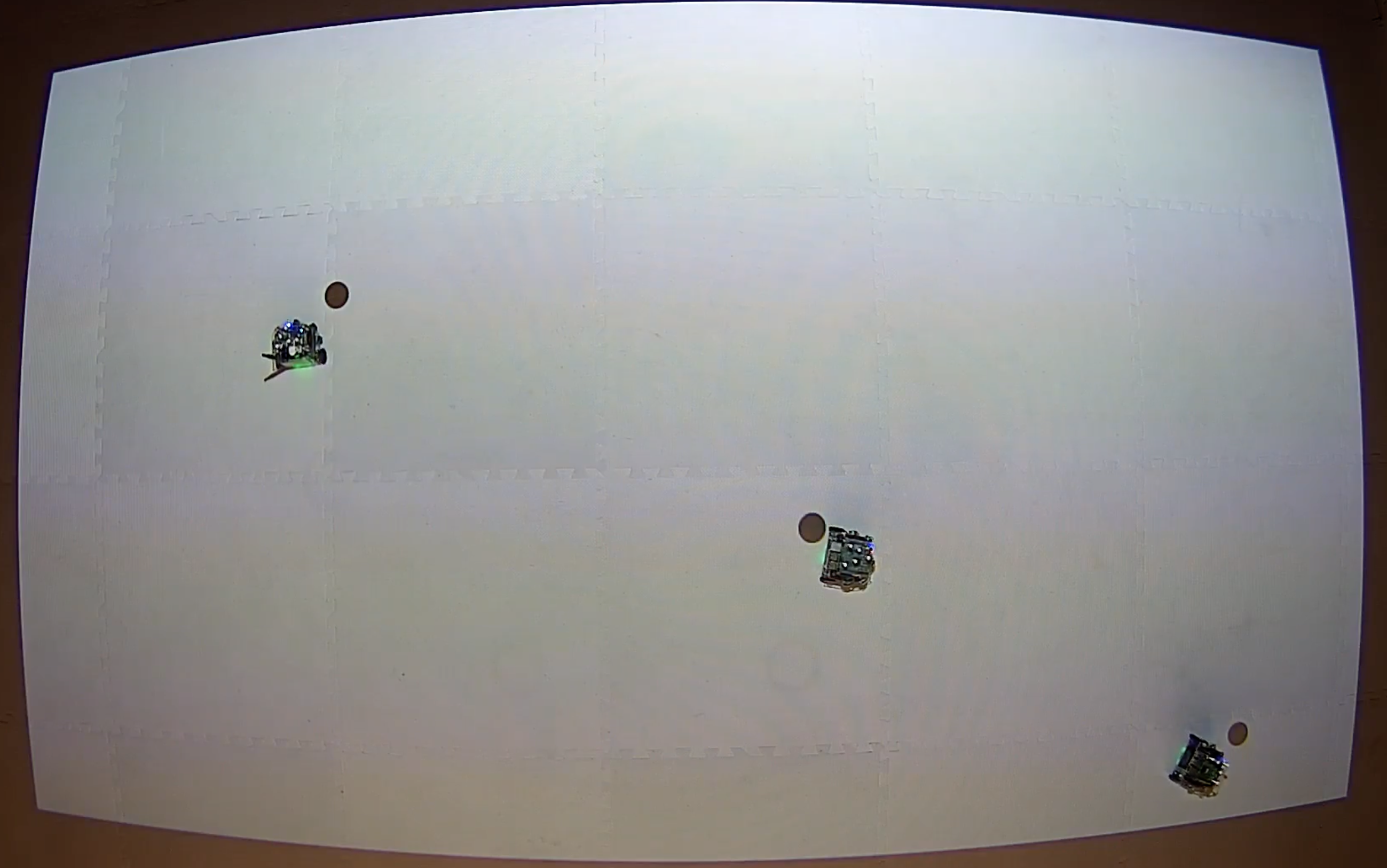}
    \end{subfigure}
    \caption{Navigation (MAPF) in simulation (left) and real-world (right).}
    \label{fig:navigation-sim-real}
\end{figure}

\subsubsection{Description}
MAPF is a classic problem in multi-robot/multi-agent systems \cite{stern2019multi, sharon2015conflict}. In this scenario, N robots must navigate to individually specified goals (black markers). Robots observe their position and the vector to their goal location. The team is rewarded for robots minimizing their distance to their respective goals. This scenario can scale to an arbitrary number of robots.

\subsubsection{Experiment Parameters}
\begin{flushitemize}
    \item Number of Robots: 3
    \item Max steps: 100
    \item Training timesteps: 3e6
    \item Controller
    \begin{flushitemize}
        \item Controller: unicycle position
        \item Barrier Fn: enabled
    \end{flushitemize}
    \item Distance Reward: -1 $\times$ euclidean distance of robots to their respective goals
    \item Violation Reward: 0 $\times$ number of collisions
\end{flushitemize}

\subsection{Predator Prey}
\begin{figure}[h]
    \centering
    \begin{subfigure}{0.48\columnwidth}
        \includegraphics[width=\columnwidth,height=0.75\columnwidth,keepaspectratio=false]{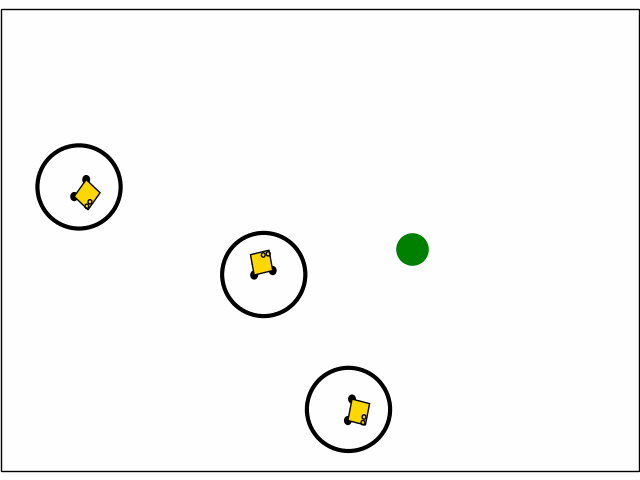}
    \end{subfigure}
    \begin{subfigure}{0.48\columnwidth}
        \includegraphics[width=\columnwidth,height=0.75\columnwidth,keepaspectratio=false]{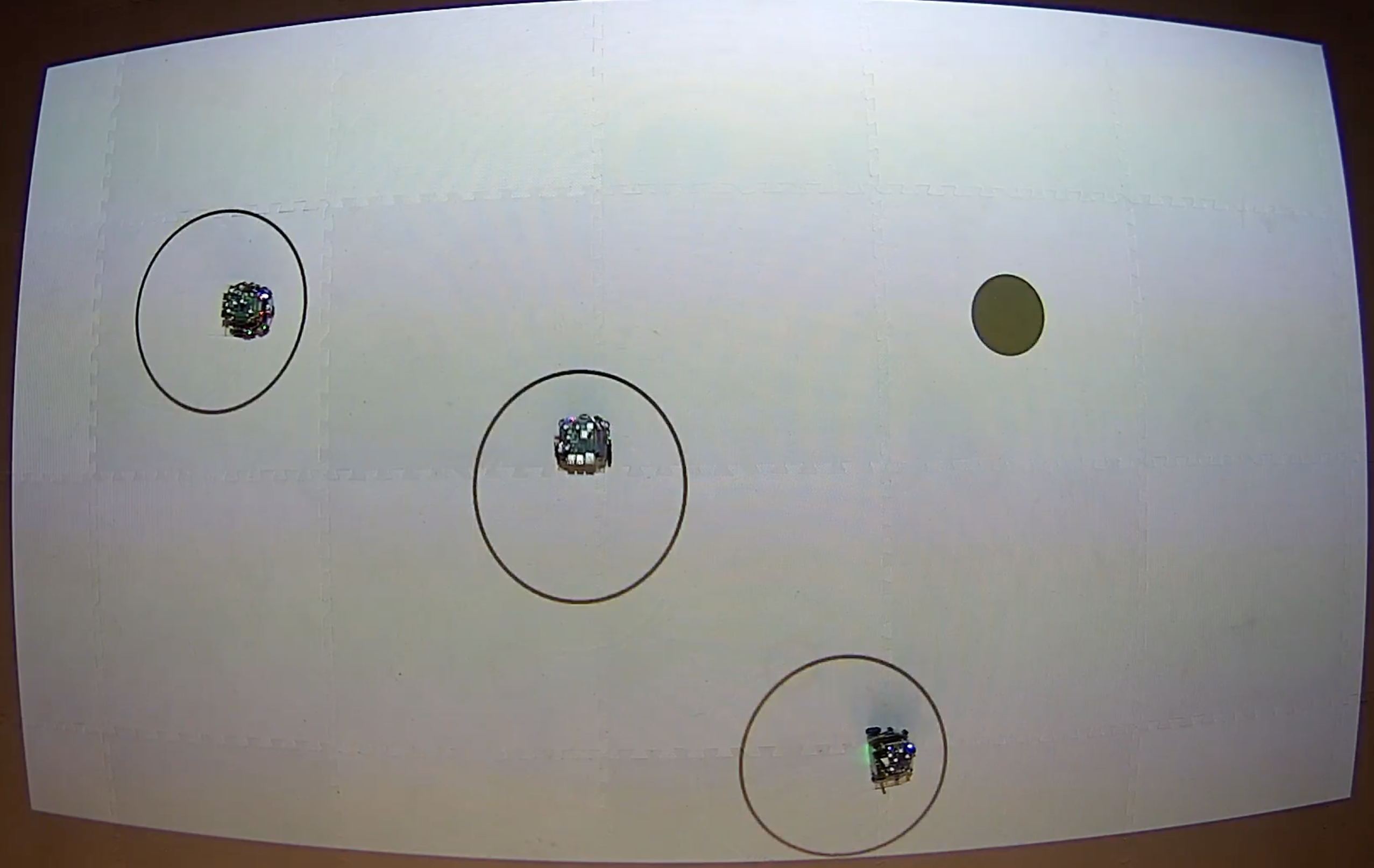}
    \end{subfigure}
    \caption{Predator Prey in simulation (left) and real-world (right).}
    \label{fig:pp-sim-real}
\end{figure}

\subsubsection{Description}
Based on a variant of the classic predator-prey scenario introduced in \cite{lowe2017multi}, in this scenario, N tagging robots must collaborate to tag a more agile prey (green marker). The prey is controlled by the heuristic proposed in \cite{peng2021facmac}, where at each step the prey maximizes its distance from the closest tagging robot. Robots observe their position, the position of other robots, and the position of the prey. Robots tag the prey by getting the prey within their tagging radius, and the prey marker will briefly turn red to indicate a successful tag. The team is rewarded for each successful tag on the prey. This scenario can scale to an arbitrary number of robots.

\subsubsection{Experiment Parameters}
\begin{flushitemize}
    \item Number of Robots: 3
    \item Max steps: 100
    \item Training timesteps: 10e6
    \item Controller
    \begin{flushitemize}
        \item Controller: unicycle position
        \item Barrier Fn: enabled
    \end{flushitemize}
    \item Tag Reward: 10 $\times$ 1 if a tag occurred, 0 otherwise
    \item Violation Reward: 0 $\times$ number of collisions
\end{flushitemize}

\subsection{Continuous-RWARE}
\begin{figure}[h]
    \centering
    \begin{subfigure}{0.48\columnwidth}
        \includegraphics[width=\columnwidth,height=0.75\columnwidth,keepaspectratio=false]{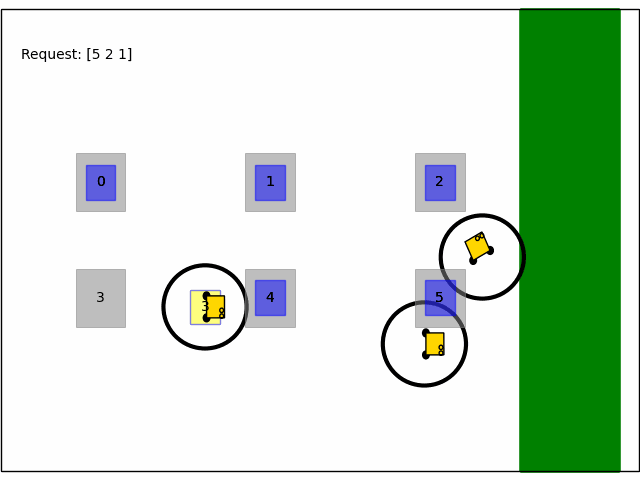}
    \end{subfigure}
    \begin{subfigure}{0.48\columnwidth}
        \includegraphics[width=\columnwidth,height=0.75\columnwidth,keepaspectratio=false]{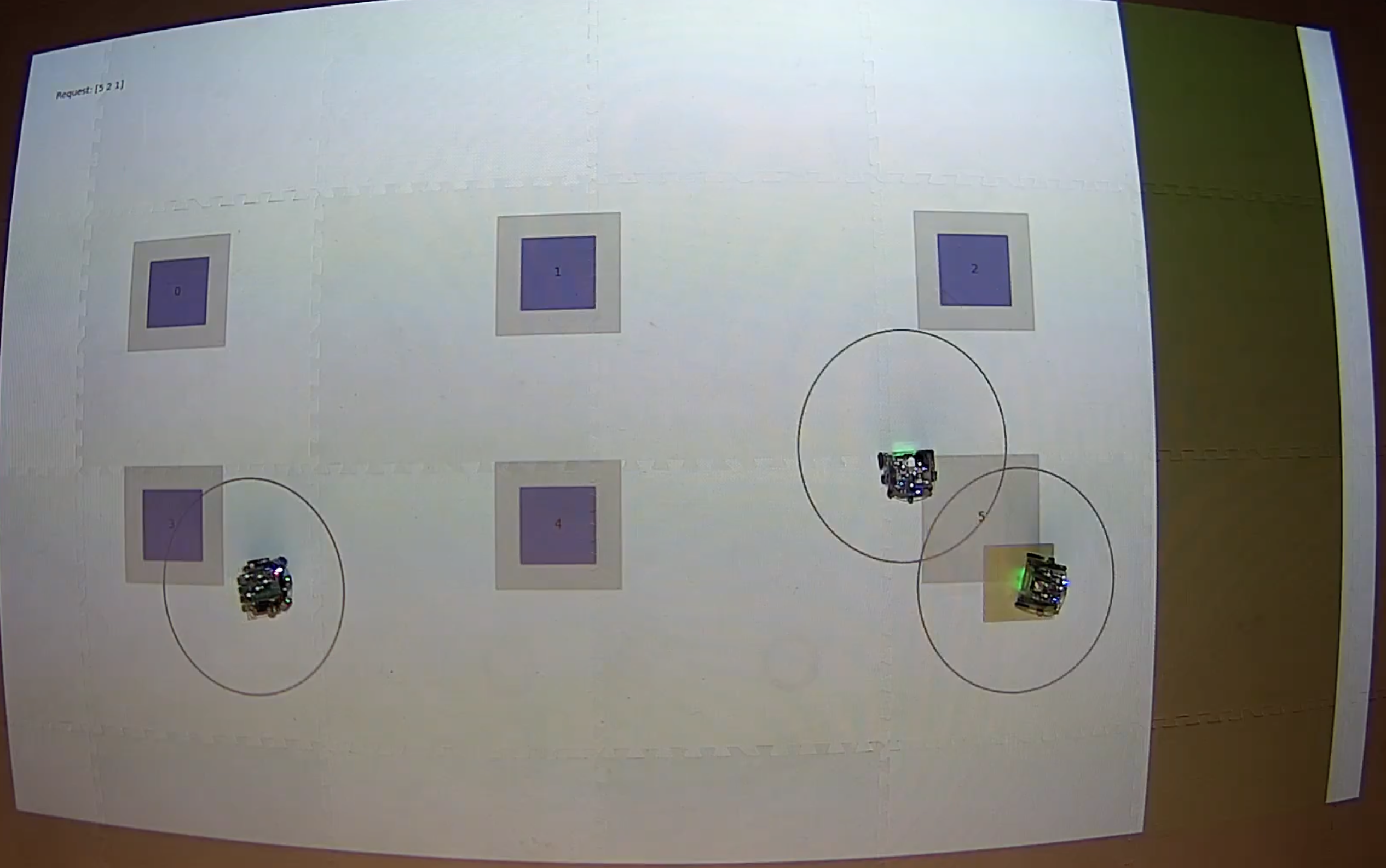}
    \end{subfigure}
    \caption{RWARE in simulation (left) and real-world (right).}
    \label{fig:rware-sim-real}
\end{figure}
\subsubsection{Description}
Inspired by grid-world RWARE \cite{rware-ref}, in this scenario, robots must collaborate to deliver the requested shelves (indicated by the top left text) to the dropoff zone (purple), where the request is updated upon each successful shelf delivery. Shelves are individually labeled and begin in staging zones (gray) arranged in a grid. Robots observe their position, the position of other robots, the position of each staging zone and the shelf occupying it, and the requested set of shelves. Robots can pickup or return shelves by entering the gray staging zone. If a robot is not carrying a shelf, it can freely move underneath shelves and through staging zones. Once a robot is carrying a shelf, it cannot move through other shelves, and shelves can only be returned at an unoccupied staging zone. A robot successfully completes a drop-off when it enters the purple zone carrying a shelf that is currently requested, at which point the corresponding entry in the request set is randomly updated to any non-requested shelf. Robots are sparsely rewarded for each successful drop-off. This scenario scales to an arbitrary number of robots and shelves.

\subsubsection{Experiment Parameters}
\begin{flushitemize}
    \item Number of Robots: 3
    \item Number of Shelves: 6
    \item Max steps: 100
    \item Training timesteps: 20e6
    \item Controller
    \begin{flushitemize}
        \item Controller: unicycle position
        \item Barrier Fn: enabled
    \end{flushitemize}
    \item Drop Off Reward: 1 $\times$ number of successful shelf drop-offs
    \item Violation Reward: 0 $\times$ number of collisions
\end{flushitemize}

\subsection{Warehouse}
\begin{figure}[h]
    \centering
    \begin{subfigure}{0.48\columnwidth}
        \includegraphics[width=\columnwidth,height=0.75\columnwidth,keepaspectratio=false]{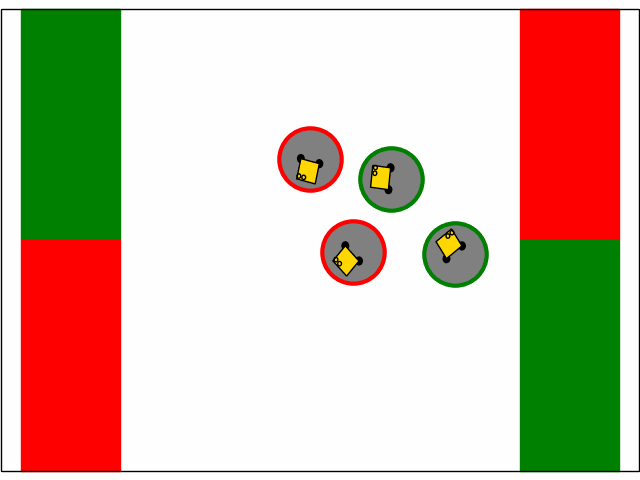}
    \end{subfigure}
    \begin{subfigure}{0.48\columnwidth}
        \includegraphics[width=\columnwidth,height=0.75\columnwidth,keepaspectratio=false]{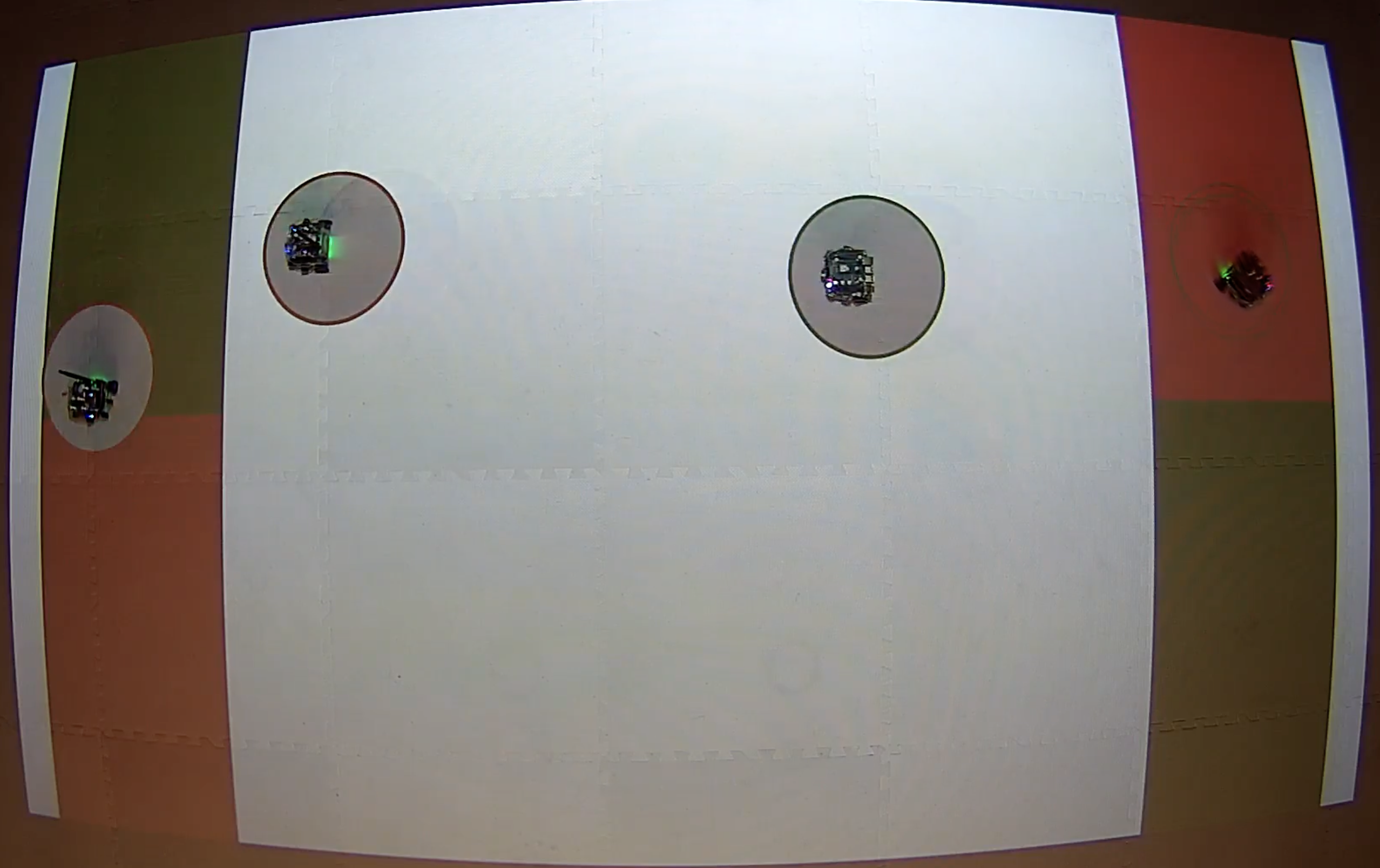}
    \end{subfigure}
    \caption{Warehouse in simulation (left) and real-world (right).}
    \label{fig:warehouse-sim-real}
\end{figure}

\subsubsection{Description}
Adopted from MARBLER, in this scenario, N red robots and M green robots must navigate to the zone with matched color on the right to pick up packages, and then deliver them to the zone with corresponding colors on the left. Robots observe their position and the position of other robots. The team is rewarded for packages loaded and packages delivered. This scenario scales to an arbitrary number of robots.

\subsubsection{Experiment Parameters}
\begin{flushitemize}
    \item Number of Robots: 4
    \item Max steps: 70
    \item Training timesteps: 3e6
    \item Controller
    \begin{flushitemize}
        \item Controller: unicycle position
        \item Barrier Fn: enabled
    \end{flushitemize}
    \item Heterogeneity
    \begin{flushitemize}
        \item Type: class
        \item Obs type: class (robots observe their own class)
        \item Values: [[1, 0], [1, 0], [0, 1], [0, 1]] (each entry indicates the type [green, red])
    \end{flushitemize}
    \item Load Reward: 1 $\times$ packages loaded
    \item Delivery Reward: 3 $\times$ packages delivered
    \item Violation Reward: 0 $\times$ number of collisions
\end{flushitemize}

\end{document}